\documentclass[preprint,12pt]{elsarticle}
\usepackage{times}
\usepackage{amsmath}
\usepackage{amssymb}
\usepackage{graphicx}
\usepackage{xcolor}
\usepackage{mathrsfs}
\usepackage{caption}

\usepackage[bottom]{footmisc}
\usepackage{bm}
\usepackage{algorithm}
\usepackage{algpseudocode}

\usepackage{mathtools}
\usepackage{subcaption}

\usepackage{float}
\usepackage{adjustbox}
\usepackage[margin=1in]{geometry}
\usepackage[small,compact]{titlesec}
\usepackage[labelfont=bf,labelsep=colon,skip=2pt]{caption} 
\usepackage[titles]{tocloft}
\usepackage{bold-extra}
\usepackage{hyperref}

\hypersetup{
    colorlinks,
    linkcolor={red!50!black},
    citecolor={blue!50!black},
    urlcolor={blue!80!black}
}

\DeclareMathOperator*{\argmax}{arg\,max}

\usepackage[left]{lineno}

\begin{document}
\begin{frontmatter}
\title{Active Learning of Model Discrepancy with Bayesian Experimental Design}

\author{Huchen Yang}
\author{Chuanqi Chen}
\author{Jin-Long Wu\corref{cor1}} \ead{jinlong.wu@wisc.edu} 
\cortext[cor1]{Corresponding author}

\address{Department of Mechanical Engineering, University of Wisconsin–Madison, Madison, WI 53706}

\begin{abstract}
Digital twins have been actively explored in many engineering applications, such as manufacturing and autonomous systems. However, model discrepancy is ubiquitous in most digital twin models and has significant impacts on the performance of using those models. In recent years, data-driven modeling techniques have been demonstrated promising in characterizing the model discrepancy in existing models, while the training data for the learning of model discrepancy is often obtained in an empirical way and an active approach of gathering informative data can potentially benefit the learning of model discrepancy. On the other hand, Bayesian experimental design (BED) provides a systematic approach to gathering the most informative data, but its performance is often negatively impacted by the model discrepancy. In this work, we build on sequential BED and propose an efficient approach to iteratively learn the model discrepancy based on the data from the BED. The performance of the proposed method is validated by a classical numerical example governed by a convection-diffusion equation, for which full BED is still feasible. The proposed method is then further studied in the same numerical example with a high-dimensional model discrepancy, which serves as a demonstration for the scenarios where full BED is not practical anymore. In this example with time-varying velocity, the network only learns the stationary discrepancy and leaves the time dependence being handled by the physics-based model itself, which helps mitigate ill-posedness of the joint learning problem. An ensemble-based approximation of information gain is further utilized to assess the data informativeness and to enhance learning model discrepancy. The results show that the proposed method is efficient and robust to the active learning of high-dimensional model discrepancy, using data suggested by the sequential BED. We also demonstrate that the proposed method is compatible with both classical numerical solvers and modern auto-differentiable solvers.
\end{abstract}


\end{frontmatter}

\section{Introduction}

Data-driven modeling techniques have been demonstrated promising in many science and engineering applications~\cite{duraisamy2019turbulence,montans_data-driven_2019,brunton2020machine}, with various types of methods being developed such as model reduction~\cite{kutz2016dynamic,peherstorfer2016data}, physics-informed machine learning~\cite{wang2017physics,wu2018physics,raissi2019physics}, operator learning~\cite{li2020fourier,lu2021learning,kochkov2021machine}, causal learning~\cite{chen2023causality,chen2023ceboosting}, and data assimilation~\cite{chen2024cgnsde, chen2025cgkn,chen2025modeling}. Considering that data informativeness serves as a critical component in all data-driven modeling techniques, an active approach to identifying the most informative data has the potential to further advance state-of-the-art data-driven modeling. Experimental design provides such a systematical approach for the active acquisition of data, which has been demonstrated effective in various types of problems, such as the estimation of some unknown model parameters~\cite{atkinson_design_1968, atkinson_optimum_1992, kirk_experimental_2009, seltman_experimental_2012}. However, classical experimental design via solving an optimization problem is limited by its propensity to fall into local optima when applied to nonlinear models and its inability to fully incorporate prior knowledge~\cite{ryan_review_2016}.

Bayesian experimental design (BED) addresses these limitations by utilizing a probabilistic framework that incorporates prior information and updates parameter distributions as new data are acquired~\cite{ryan_review_2016, jones_bayes_2016}. This approach not only enhances the robustness of the design and data acquisition process but also provides a more flexible and adaptive framework that can efficiently handle complex, nonlinear models \cite{huan_simulation-based_2013,rainforth_modern_2023}. Foundational work \cite{raiffa_applied_2000,degroot_optimal_2005} introduced Bayesian decision theory into the field of experimental design, emphasizing the importance of selecting a utility function that appropriately reflects the purpose of the experiment, which is critical for identifying optimal designs \cite{lindley_bayesian_1972}, and various popular criteria used to find optimal designs were summarized in~\cite{chaloner_bayesian_1995}. The growing popularity of sequential or adaptive designs in Bayesian design literature highlights their flexibility and efficiency over static designs \cite{whitehead_bayesian_1995,cavagnaro_adaptive_2010,drovandi_sequential_2013}. Numerous novel strategies and techniques have been developed to overcome the computational challenges for Bayesian optimal design problems~\cite{ivanova_implicit_2021,foster_deep_2021,huan_gradient-based_2014,shen_bayesian_2023}, and a comprehensive review of BED for its formulations and computations can be found in~\cite{huan2024optimal}.

Sequential Bayesian experimental design (sBED) has emerged as a significant methodology in experimental design, leveraging the results of preceding experiments to inform subsequent decisions \cite{muller_simulation-based_2007,von_toussaint_bayesian_2011,huan_numerical_2015}. It incorporates adaptive decision-making and feedback, making it highly effective in iterative experimentation. Among sBED methods, there are two main streams: greedy (or myopic) identification of the optimal design for the current situation using only the information obtained so far \cite{box_sequential_1992, drovandi_sequential_2014,kleinegesse_sequential_2020}, while policy-based approaches considers all future experiments when making each design decision  \cite{shen_bayesian_2023,huan_sequential_2016}. Consequently, the policy-based approaches are typically finite-horizon, which requires the total number of experiments to be predetermined. In the context of active online learning, where information and system states are continuously updated and the scale of the experiment cannot be pre-specified, the greedy approach tends to be more flexible.

However, nearly all BED methods suffer from model discrepancy, a phenomenon that is prevalent in real-world scenarios \cite{kennedy_bayesian_2001}. The model discrepancy primarily introduces errors in the likelihood calculation of BED, leading to the following consequences: (i) introduction of bias in parameter estimation; (ii) continuous generation of similar designs and low-quality datasets, which present significant challenges in BED practice \cite{grunwald_inconsistency_2017,brynjarsdottir_learning_2014,catanach_metrics_2023}. In the traditional BED framework, two primary categories of approaches are employed to address the issue of model discrepancy. The first category involves augmenting the model with correction terms that contain some unknown parameters to be inferred \cite{feng_optimal_2015, surer_sequential_2024}. However, this approach substantially increases the dimensionality of the parameter space, especially when large neural networks are employed to construct those correction terms, thereby introducing significant computational challenges. The second category entails proposing a set of candidate models and either evaluating the robustness of the discrepancy-mitigating designs using new criteria \cite{catanach_metrics_2023}, or selecting the most plausible model through model discrimination \cite{donckels_anticipatory_2009, streif_optimal_2014}. Model discrimination can also be integrated into the parameter inference process, resulting in a multi-objective optimization problem for experimental design. This category of methods is constrained by the necessity of ensuring that the candidate model space includes the true model, a requirement that is often challenging and costly to fulfill.

To efficiently handle the model discrepancy in the context of BED, we propose a hybrid framework that integrates online learning for correcting the model discrepancy with sequential BED methods. An optimization-based method is adopted to gradually update a neural-network-based correction term that characterizes the model discrepancy, using the online data provided by each stage of sequential BED. It is worth noting that the neural-network-based correction term can potentially demand a high-dimensional parameter space, and the proposed approach avoids dealing with high-dimensional distributions of the unknown parameters, which poses a key challenge to Bayesian inference. At each stage of sequential BED, the proposed hybrid framework first employs BED for the parameters of the existing model and then performs optimization for the neural-network-based correction term. On one hand, each stage of sequential BED identifies optimal designs and the corresponding data to update the correction term. On the other hand, the updated correction term at each stage gradually mitigates the model discrepancy in BED.

To ensure the informativeness of data for correcting the model discrepancy, we introduce an ensemble-based indicator that approximates the information gain from a given dataset for the calibration of the model discrepancy, based on the ensemble Kalman inversion (EKI)~\cite{iglesias_ensemble_2013, kovachki_ensemble_2019}. More specifically, we propose an efficient and robust indicator to assess the data informativeness under the Gaussian assumption, which allows us to use the initial and updated ensembles from EKI to approximate the information gain for a candidate design and the corresponding measurement data. This process is more computationally feasible than those more accurate Bayesian approaches such as Markov chain Monte Carlo (MCMC), while still effectively quantifying the data informativeness and avoiding the calibration of model discrepancy on less informative or even misleading data, which could lead to a biased solution.

The key highlights of our work are as follows.
\begin{itemize}
    \item We propose a hybrid framework for efficient nonlinear BED and validate its performance in a numerical example with a low-dimensional parametric model error, for which full BED is still feasible.
    \item We further demonstrate that the hybrid framework can efficiently deal with structural errors characterized by high-dimensional parameters in the existing model, for which fully BED becomes impractical.
    \item We establish an ensemble-based indicator to approximate the information gain and improve the efficiency and robustness of the hybrid framework in the active learning of model discrepancy. 
\end{itemize}

This paper is organized as follows. Section \ref{sec: Methodology} introduces the general formulation of our method. Section \ref{sec: Numerical Results} provides a comprehensive study of a contaminant source inversion problem in a convection-diffusion field to demonstrate the performance of the proposed method. Section~\ref{sec:discussions} presents discussions of potential future extensions. Finally, Section \ref{Conclusion} concludes the paper.

\section{Methodology}\label{sec: Methodology}
The model discrepancy is ubiquitous in the modeling and simulation of complex dynamical systems~\cite{wu2024learning} and poses a key challenge to the performance of Bayesian optimal experimental design. Unlike the unknown parameters of an existing physics-based model that are often in a low-dimensional vector space, the model discrepancy is in infinite-dimensional function space, for which the approximation approaches (e.g., based on neural networks) can involve unknown coefficients in a high-dimensional vector space and thus make full Bayesian approach such as MCMC practically infeasible. The general goal of this work is to leverage the optimal data identified initially for interested physical parameters by the standard BED via an ensemble-based method and to efficiently learn the unknown coefficients of the neural network that approximates the model discrepancy. 

In this work, the true system can be written in a general form:
\begin{equation}
\label{eq:true_system}
\begin{aligned}
    \frac{\partial \mathbf{u}}{\partial t} = \mathcal{G}^\dagger(\mathbf{u}; \boldsymbol{\theta}^\dagger),
\end{aligned}
\end{equation}
where the system state $\mathbf{u}(\mathbf{z}, t)$ is often a spatiotemporal field for many engineering applications, and $\mathcal{G}^\dagger$ denotes the true dynamics, which may include differential and/or integral operators. If the detailed form of true dynamics is known, the BED can be used straightforwardly to estimate the unknown parameters $\boldsymbol{\theta}^\dagger$. In practice, the form of true dynamics is often unknown for many complex dynamical systems. Assuming that a physics-based model $\mathcal{G}(\mathbf{u};\boldsymbol{\theta}_\mathcal{G})$ exists to approximate the true dynamics $\mathcal{G}^\dagger$, we focus on the learning of model discrepancy $\mathcal{G}^\dagger - \mathcal{G}$, which is characterized by a neural network $\textrm{NN}(\mathbf{u};\boldsymbol{\theta}_\textrm{NN})$ and leads to the modeled system:

\begin{equation}
\label{eq:modeled_system}
    \frac{\partial \mathbf{u}}{\partial t}=\mathcal{G}(\mathbf{u};\boldsymbol{\theta}_\mathcal{G})+\textrm{NN}(\mathbf{u};\boldsymbol{\theta}_\textrm{NN}).
\end{equation}

Using a neural-network-based model to characterize the model discrepancy has recently been explored with different choices of neural networks such as neural operators~\cite{chen2025neural} and diffusion models~\cite{dong2024data, dong2025stochastic}, and with various types of research context such as ensemble data assimilation~\cite{luo_accounting_2021}. Although ill-posedness could generally be an issue for the joint learning of physics-based model and neural-network-based discrepancy, systematically preventing the neural network from ‘explaining away’ physically prescribed terms is not the main focus of this work, and the numerical examples studied in this work do not noticeably suffer from such an issue.

In the context of BED, the design $\mathbf{d}$ corresponds to the spatiotemporal coordinate $(\mathbf{z}, t)$. The measurement $\mathbf{y}$ is the observation of variable $\mathbf{u}$ that represents the state of the physical system at the design $\mathbf{d}$, i.e., $\mathbf{y} = \mathbf{u}(\mathbf{d}) + \boldsymbol\eta$, where $\boldsymbol\eta \sim \mathcal{N}(\boldsymbol{0}, \boldsymbol\Gamma)$ represents the measurement noise which is assumed to be Gaussian in this work when observing the state from the true system. More generally, the proposed framework can be extended to nonlinear observation systems, provided that the corresponding likelihood function can be explicitly defined based on the noise model.

\subsection{Bayesian Experimental Design}
\label{sec:OED}

The Bayesian experimental design~\cite{rainforth_modern_2023,huan2024optimal} provides a general framework to systematically seek the optimal design by solving the optimization problem: 
\begin{equation}
\label{eq:oed_opt}
    \begin{aligned}
        \mathbf{d}^* &= \argmax_{\mathbf{d}\in\mathcal{D}} \mathbb{E}[U(\boldsymbol{\theta},\mathbf{y},\mathbf{d})]\\
        &= \argmax_{\mathbf{d}\in\mathcal{D}} \int_\mathcal{Y} \int_{\boldsymbol{\Theta}} U(\boldsymbol{\theta},\mathbf{y},\mathbf{d}) p(\boldsymbol{\theta},\mathbf{y}|\mathbf{d}) \mathrm{d}\boldsymbol{\theta}\mathrm{d}\mathbf{y}\\
        &= \argmax_{\mathbf{d}\in\mathcal{D}} \int_\mathcal{Y} \int_{\boldsymbol{\Theta}} U(\boldsymbol{\theta},\mathbf{y}, \mathbf{d}) p(\mathbf{y}|\boldsymbol{\theta},\mathbf{d})p(\boldsymbol{\theta}|\mathbf{d}) \mathrm{d}\boldsymbol{\theta}\mathrm{d}\mathbf{y},
    \end{aligned}
\end{equation}
where $\mathbf{y} \in \mathcal{Y} \subseteq \mathbb{R}^{d_{\mathbf{y}}}$ is data from the design $\mathbf{d} \in \mathcal{D}\subseteq \mathbb{R}^{d_{\mathbf{d}}}$, $\boldsymbol{\theta} \in \boldsymbol{\Theta}\subseteq \mathbb{R}^{\boldsymbol{\theta}}$ denotes the target parameters, and $U$ is the utility function taking the inputs of $\mathbf{y}$ and $\boldsymbol{\theta}$ and returning a real value. The optimal design $\mathbf{d}^*$ is obtained by maximizing the expected utility function $\mathbb{E}[U(\boldsymbol{\theta},\mathbf{y},\mathbf{d})]$ over the design space $\mathcal{D}$. The term $p(\boldsymbol{\theta},\mathbf{y}|\mathbf{d})$ is the joint conditional distribution of data and parameters. $p(\mathbf{y}|\boldsymbol{\theta},\mathbf{d})$ is the likelihood. One thing that should be noted is that $p(\boldsymbol{\theta}|\mathbf{d})$ is the prior distribution of parameters, which is often assumed to be independent of the design $\mathbf{d}$: $p(\boldsymbol{\theta}|\mathbf{d}) = p(\boldsymbol{\theta})$.

The utility function $U(\boldsymbol{\theta},\mathbf{y},\mathbf{d})$ can be regarded as the reward obtained from the data $\mathbf{y}$ for the corresponding design $\mathbf{d}$. This work employs Kullback-Leibler divergence between the posterior and the prior distribution of parameters $\boldsymbol{\theta}$, which is a widely used utility function in BED~\cite{shen_bayesian_2023} and has an information-theoretic interpretation\cite{ryan_review_2016, chaloner_bayesian_1995, rainforth_modern_2023}:
\begin{equation}
    \begin{aligned}
        U(\boldsymbol{\theta},\mathbf{y},\mathbf{d}) &= D_{\textrm{KL}} \bigl(p(\boldsymbol\theta|\mathbf{y},\mathbf{d})\| p(\boldsymbol\theta)\bigr)\\
        &=\mathbb{E}_{\boldsymbol\theta|\mathbf{d},\mathbf{y}}(\log p(\boldsymbol{\theta}|\mathbf{y},\mathbf{d}) - \log p(\boldsymbol{\theta}))\\
        &=\int_{\boldsymbol{\Theta}} p(\boldsymbol{\theta}|\mathbf{y},\mathbf{d}) \log(\frac{p(\boldsymbol{\theta}|\mathbf{y},\mathbf{d})}{p(\boldsymbol{\theta})})\mathrm{d}\boldsymbol{\theta},
    \end{aligned}
    \label{eq: kld utility}
\end{equation}
where $p(\boldsymbol\theta|\mathbf{y},\mathbf{d})$ is the posterior distribution of $\boldsymbol\theta$ given a design $\mathbf{d}$ and the data $\mathbf{y}$. Note that data $\mathbf{y}$ could be either from the actual experimental measurement or the numerical model simulation. To identify an optimal design without performing any experiments, expected information gain (EIG,~\cite{lindley_measure_1956}) is employed to integrate the information gain over all possible predicted data:
\begin{equation}
\label{eq:EIG}
    \begin{aligned}
        \mathbb{E}[U(\boldsymbol{\theta},\mathbf{y},\mathbf{d})] &= \text{EIG} (\mathbf{d}) \\
        &=\mathbb{E}_{\mathbf{y} | \mathbf{d}} [D_{\textrm{KL}}(p(\boldsymbol\theta|\mathbf{y},\mathbf{d})\| p(\boldsymbol\theta))]\\
  &=\int_{\mathcal{Y}}D_{\textrm{KL}}(p(\boldsymbol\theta|\mathbf{y},\mathbf{d})\| p(\boldsymbol\theta)) p(\mathbf{y}|\mathbf{d})\mathrm{d}\mathbf{y}\\      
    \end{aligned}
\end{equation}
where $p(\mathbf{y}|\mathbf{d}):=\mathbb{E}_{\boldsymbol{\theta}}[p(\mathbf{y}|\boldsymbol{\theta},\mathbf{d})]$ is the distribution of the predicted data among all possible parameter values given a certain design. It should be noted that designs given by maximizing EIG solely depend on prior parameter belief \cite{zemplenyi_bayesian_2021} and an appropriate model \cite{rainforth_modern_2023}. Although this utility still works for the weak model discrepancy cases \cite{feng_optimal_2015}, the stronger model discrepancy will likely lead to correlated biases in likelihoods and non-negligible impacts on the EIG, thus resulting in designs of poor quality \cite{brynjarsdottir_learning_2014, catanach_metrics_2023}. This could be alleviated by introducing some information from the true system, i.e. actual measurements, as preliminary data~\cite{waldron_closed-loop_2019,wang_measure_2024}.

Beyond each belief update, we mainly focus on the scenario of sequential BED \cite{shen_bayesian_2023}, which requires sequentially performing experiments to update the design. With the aggregated history information of optimal design and the corresponding measured data denoted as $I_i = [\mathbf{y}_1, \mathbf{d}_1, \mathbf{y}_2, \mathbf{d}_2, \allowbreak \cdots, \mathbf{y}_i, \mathbf{d}_i]$, the Bayesian updating process in the sequential BED at $i$-th stage can be written as:

\begin{equation}p(\boldsymbol\theta|I_{i})=\frac{p(\mathbf{y}_{i}|\boldsymbol\theta,\mathbf{d}_{i}, I_{i-1})p(\boldsymbol\theta|I_{i-1})}{p(\mathbf{y}_{i}|\mathbf{d}_{i})},
\end{equation}
where $p(\boldsymbol\theta|I_i)$ is the posterior distribution at $i$-th stage and $p(\mathbf{y}_{i}|\boldsymbol\theta,\mathbf{d}_{i}, I_{i-1})$ is the likelihood based on data $\mathbf{y}_{i}$, design $\mathbf{d}_{i}$ and aggregated information $I_{i-1}$. If using $p(\boldsymbol\theta|I_{i-1})$ and $p(\boldsymbol\theta|I_{i})$ to replace the prior distribution $p(\boldsymbol\theta)$ and the posterior one $p(\boldsymbol\theta|\mathbf{y},\mathbf{d})$, we can define the EIG at $i$-th stage and denote it as $\mathbb{E}[U(\boldsymbol{\theta},\mathbf{y},\mathbf{d})]_i$. In this work, a greedy strategy that maximizes $\mathbb{E}[U(\boldsymbol{\theta},\mathbf{y},\mathbf{d})]$ at every stage has been taken:
\begin{equation}
    \label{eq:optimal_design_i}
    \mathbf{d}_{i}= \argmax_{\mathbf{d} \in \mathcal{D}}\mathbb{E}[U(\boldsymbol{\theta},\mathbf{y},\mathbf{d})]_i.
\end{equation}

In this work, the optimization problem is solved using a gradient-based approach, for which the gradient can be approximated via Infinitesimal Perturbation Analysis (IPA)~\cite{huan_gradient-based_2014}. In practice, it can also be computed using automatic differentiation. For the high-dimensional designs, comprehensive reviews of methods for solving design optimization problems can be found in~\cite{rainforth_modern_2023, huan2024optimal}.

\subsection{Model Discrepancy Correction}
\label{sec: Model error correction}

The presence of model discrepancy, i.e., the difference between $\mathcal{G}^\dagger$ in Eq.~\eqref{eq:true_system} and $\mathcal{G}$ in Eq.~\eqref{eq:modeled_system}, would inevitably introduce bias in the estimated parameters $\boldsymbol{\theta}_\mathcal{G}$ and thus motivate the study of quantifying the model discrepancy $\mathcal{G}^\dagger-\mathcal{G}$ for BED~\cite{feng_optimal_2015}. In this work, we investigate the use of neural networks to characterize the model discrepancy as shown in Eq.~\eqref{eq:modeled_system}, for which the key challenge is that the unknown parameters $\boldsymbol{\theta}_\text{NN}$ are often high-dimensional to fully exploit the representation power of neural networks and thus pose a computational challenge to Bayesian inference. Therefore, a common practice is to calibrate the model discrepancy offline with empirically chosen data. However, offline calibration relies heavily on the quality and availability of pre-selected data. In this work, we focus on the active online learning of model discrepancy parameters $\boldsymbol{\theta}_\text{NN}$, which exploits the optimal data obtained via sequential BED to estimate physics-based model parameters $\boldsymbol{\theta}_\mathcal{G}$ in Eq.~\eqref{eq:modeled_system}. 

Considering that $\boldsymbol{\theta}_\mathcal{G}$ is correlated with the model discrepancy, the optimal data for $\boldsymbol{\theta}_\mathcal{G}$ obtained via sequential BED is likely to be still informative in the calibration of the model discrepancy, which can be efficiently handled by solving an optimization problem. Therefore, this work employs an iterative approach to actively gather informative data and to estimate the unknown parameters $\boldsymbol{\theta}_\mathcal{G}$ and $\boldsymbol{\theta}_\text{NN}$ in Eq.~\eqref{eq:modeled_system}. 

More specifically, the iterative approach at $i$-th stage starts with solving a BED problem for $\boldsymbol{\theta}_\mathcal{G}$ based on the current estimation of $\boldsymbol{\theta}_\text{NN}$ to obtain the optimal design $\mathbf{d}$ in Eq.~\eqref{eq:optimal_design_i}. With the fixed design $\mathbf{d}$ and the neural network parameters $\boldsymbol{\theta}_\text{NN}$, we further obtain the maximum a posteriori (MAP) point $\boldsymbol\theta_\mathcal{G}^*$:

\begin{equation}
    \label{eq:selet 1 theta}
    \boldsymbol\theta_\mathcal{G}^*=\argmax_{\boldsymbol\theta_\mathcal{G}}
 \{ p(\mathbf{y}|\boldsymbol\theta_\mathcal{G},\mathbf{d},\boldsymbol\theta_\text{NN})p(\boldsymbol\theta_\mathcal{G}|\boldsymbol\theta_\text{NN})  \}.
\end{equation}
where $p(\boldsymbol\theta_\mathcal{G}|\boldsymbol\theta_\text{NN}) $ represents the prior distribution with $\mathbf{d}$ hiden since prior knowledge is independant with experimental design.

With the MAP estimation of $\boldsymbol\theta_\mathcal{G}^*$ further being fixed, the objective function $L(\boldsymbol\theta_\text{NN};\ \boldsymbol\theta_\mathcal{G}^*,\mathbf{y},\mathbf{d})$ is defined by the likelihood function $p(\mathbf{y}|\boldsymbol\theta_\mathcal{G}^*,\mathbf{d},\boldsymbol\theta_\text{NN})$ , and the updated estimation of $\boldsymbol\theta_\text{NN}$ can be obtained via:

\begin{equation}
    \label{eq:optimal_theta_NN}
    \boldsymbol\theta_\text{NN}^*=\argmax_{\boldsymbol\theta_\text{NN}} L(\boldsymbol\theta_\text{NN};\boldsymbol\theta_\mathcal{G}^*,\mathbf{y},\mathbf{d}).
\end{equation}

\begin{figure}[H]
    \centering
    \includegraphics[width=\linewidth]{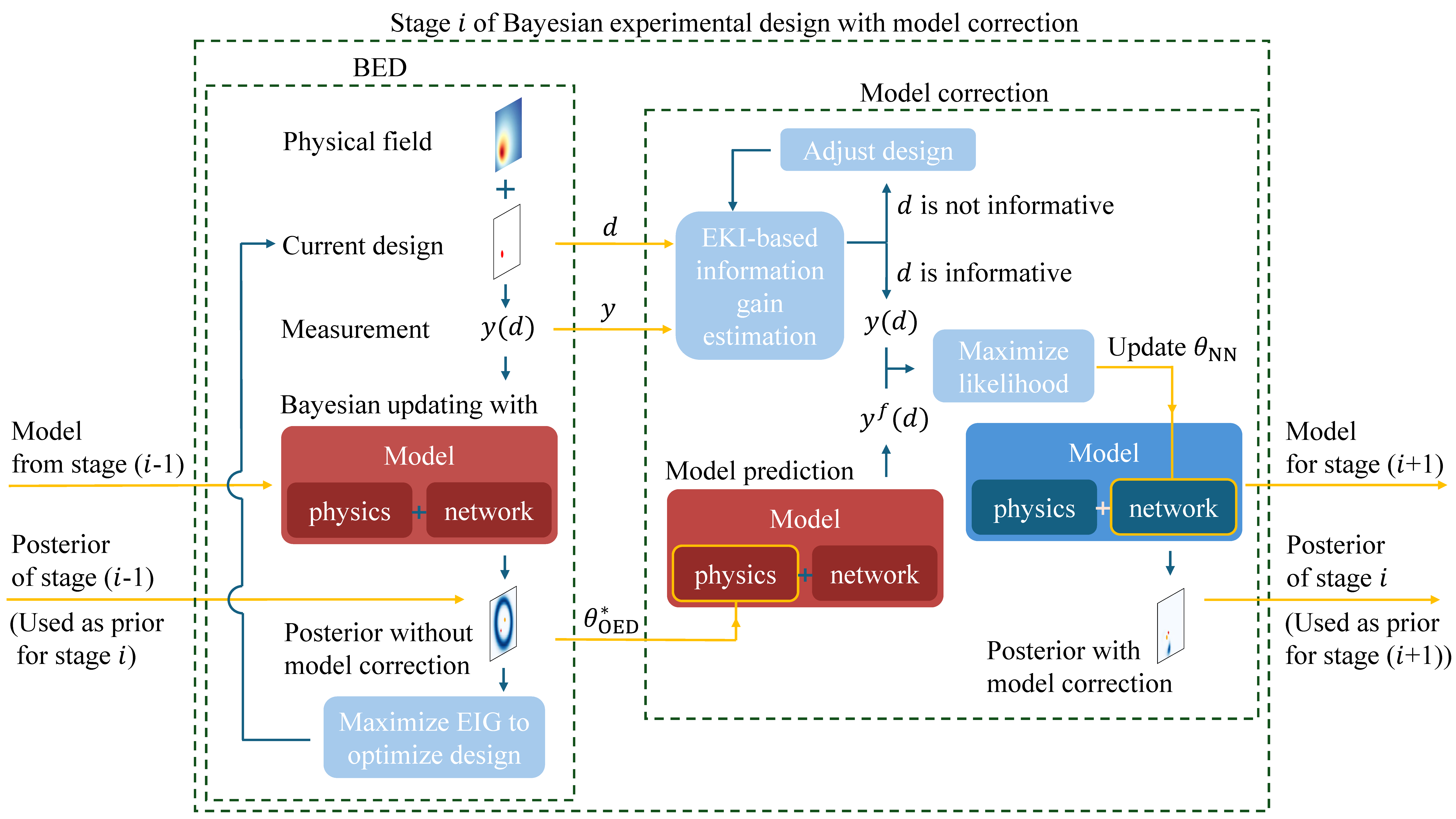}
    \caption{Schematic diagram of the hybrid framework for sequential BED and active learning of model discrepancy. The prior and posterior distributions refer to the probability distribution of the parameter $\theta_\mathcal{G}$ in Eq.~\eqref{eq:modeled_system}.}
    \label{fig:graphic_abstract}
\end{figure}

We omit the stage index $i$ in this section for simplicity in notation. In summary, there are three key steps at each stage of the hybrid framework illustrated in Fig.~\ref{fig:graphic_abstract}: (i) performing BED to obtain optimal design $\mathbf{d}$, (ii) obtaining the MAP estimation $\boldsymbol\theta_\mathcal{G}^*$, and (iii) updating the estimation of $\boldsymbol{\theta}_{\text{NN}}$. A sketch of proof for the existence of a globally optimal solution is presented in~\ref{appendx proof for obj in model correction}. Here, the global optimum refers to the solution that the MAP estimate of $\boldsymbol{\theta}_\mathcal{G}$ aligns with the ground truth and the neural network perfectly recovers the discrepancy function. It is worth noting that gradient descent optimization methods would not always guarantee such a global optimal solution.

This iterative procedure exploits a distinction between the two types of parameters. We treat $\boldsymbol{\theta}_{\mathcal{G}}$ as a low-dimensional random vector inferred through Bayesian approaches. In contrast, the neural network parameters $\boldsymbol{\theta}_{\text{NN}}$ are optimized via gradient-based methods due to their high dimensionality, which makes a full Bayesian viewpoint computationally prohibitive. The above updating strategy resembles the Expectation-Maximization (EM) algorithm, in which one parameter set takes the expectation to facilitate the optimization of the other parameter set.

In practice, a single MAP estimation via Eq. \eqref{eq:optimal_theta_NN} can sometimes lead to unsatisfactory performance. At intermediate stages of the sequential design, a limited amount of data can potentially lead to a posterior landscape with multiple, potentially misleading, local maxima. Any single one of these maxima could bias the algorithm and hinder subsequent decisions. To enhance the robustness of the proposed algorithm, we therefore seek to identify multiple local maxima from the posterior distribution of $\boldsymbol\theta_\mathcal{G}$:
\begin{equation}
    \{\boldsymbol\theta_\mathcal{G}^{(1)},\boldsymbol\theta_\mathcal{G}^{(2)},...,\boldsymbol\theta_\mathcal{G}^{(m)}\}=\operatorname{arg\,top}_m p(\mathbf{y}|\boldsymbol\theta_\mathcal{G},\mathbf{d},\boldsymbol\theta_\text{NN})p(\boldsymbol\theta_\mathcal{G}|\boldsymbol\theta_\text{NN}),\label{eq:selet m theta}
\end{equation}
where $\operatorname{arg\,top}_m$ denotes the process that identifies the set of parameters corresponding to the top $m$ largest values of a function. This can be approximated by solving the MAP problem in Eq.~\eqref{eq:selet 1 theta} $m$ times with randomly chosen initial points. The $\boldsymbol\theta_\mathcal{G}^*$ in Eq.~\eqref{eq:optimal_theta_NN} is then approximated by those samples (more related to posterior mean) :
\begin{equation}
    \boldsymbol\theta_\mathcal{G}^* \approx \frac{1}{m}\sum_{i=1}^m \boldsymbol\theta_\mathcal{G}^{(i)}.
\end{equation}
The primary purpose of this averaging is to produce a robust point estimate $\boldsymbol\theta_\mathcal{G}^*$ for the subsequent neural network update. While it is not guaranteed to be a local maximum of the posterior, numerical results suggest that the gradient it provides still benefits updating the neural network parameters. More specifically, the gradient of the objective function in Eq.~\eqref{eq:optimal_theta_NN} evaluated at the averaged point provides a descent direction that is balanced by the influence of all identified peaks. This mechanism prevents the network from overfitting to a single, potentially misleading, mode during the learning process. The limitations of this averaging strategy, particularly in cases where the ground truth posterior is genuinely multimodal, are further discussed in Section~\ref{sec:discussions}.

The optimization problem in Eq.~\eqref{eq:optimal_theta_NN} can be efficiently solved via gradient descent methods. The gradient of the objective function in Eq.~\eqref{eq:optimal_theta_NN} with respect to the neural network parameters is written as:

\begin{equation}
    \frac{\mathrm{d} \text{Obj}}{\mathrm{d} \boldsymbol\theta_\text{NN}} = \frac{\partial \text{Obj}(\mathbf{u})}{\partial \mathbf{u}} \frac{\mathrm{d} \mathbf{u}}{\mathrm{d} \boldsymbol\theta_\text{NN}}.
\end{equation}

It should be noted that $\mathrm{d} \mathbf{u}/\mathrm{d} \boldsymbol\theta_\text{NN}$ is often expensive to evaluate directly when $\mathbf{u}$ and $\boldsymbol\theta_\text{NN}$ are both high-dimensional. Therefore, efficient methods have been developed to obtain $\mathrm{d} \text{Obj}/\mathrm{d} \boldsymbol\theta_\text{NN}$ (i.e., $\partial L / \partial \boldsymbol\theta_\text{NN}$) such as automatic differentiation~\cite{griewank_automatic_1988,frostig_compiling_nodate} or adjoint methods~\cite{hilbert_methods_1985,givoli_tutorial_2021}. 

As automatic differentiation is essentially a discretized version of the adjoint method, we only briefly introduce the application of the adjoint method here. The brief introduction below is based on data at a given time step $t_1$, while derivations and extensions for handling time-integrated objective functions and data at multiple time steps can be found in~\cite{cao_adjoint_2002, givoli_tutorial_2021}.

To avoid directly calculating $\mathrm{d} \mathbf{u}/\mathrm{d} \boldsymbol\theta_\text{NN}$, an adjoint state $\lambda(t)$ is defined such that $\lambda(t)^\top=\partial L / \partial \mathbf{u}(t)$. For the modeled system in Eq.~\eqref{eq:modeled_system}, the adjoint state in time range $[t_0,t_1]$ can be obtained by backwardly solving:

\begin{equation}
\left\{
    \begin{array}{l}
        \begin{aligned}
            \frac{\mathrm{d}\lambda(t)}{\mathrm{d}t} &= -\left( \frac{\partial (\mathcal{G}+\text{NN})}{\partial \mathbf{u}} \right)^\top \lambda(t)\\
            \lambda^\top(t_1) &= \frac{\partial L}{\partial \mathbf{u}(t_1)}
        \end{aligned}
    \end{array}
\right.,
\end{equation}
where $\partial \text{NN}/\partial \mathbf{u}$ can be obtained directly via the backpropagation of the neural network $\text{NN}$ and $\partial \mathcal{G}/\partial \mathbf{u}$ is the Jacobian of the physics-based model $\mathcal{G}$. In practice, the Jacobian can be derived analytically or approximated by the coefficient matrix from the numerical solver, which often formulates a linear system. With the adjoint state obtained, the gradient $\mathrm{d} L /\mathrm{d} \boldsymbol\theta_\text{NN}$ can then be calculated as follows:

\begin{equation}
    \label{eq:adjoint_grad}
    \begin{aligned}
        \frac{\mathrm{d} L}{\mathrm{d} \boldsymbol\theta_\text{NN}}&=\int_{t_0}^{t_1} \lambda^\top(t)\frac{\partial \left(\mathcal{G}+\text{NN}\right)}{\partial \boldsymbol\theta_\text{NN}} dt\\
        &=\underbrace{
    \int_{t_0}^{t_1} \lambda^\top(t)\frac{\partial \mathcal{G}}{\partial \boldsymbol\theta_\text{NN}} dt
}_{\mathclap{=0}}+\int_{t_0}^{t_1} \lambda^\top(t)\frac{\partial \text{NN}}{\partial \boldsymbol\theta_\text{NN}} dt,
    \end{aligned}
\end{equation}
where the first term at the right-hand side has to be zero, as the physics-based model $\mathcal{G}$ does not depend on $\theta_\text{NN}$.

\subsection{Ensemble-based Indicator for Data Informativeness} 
\label{sec:ensemble_indicator}

The key idea behind the iterative approach in Section~\ref{sec: Model error correction} for the calibration of unknown parameters $\boldsymbol\theta_\text{NN}$ is that $\boldsymbol{\theta}_\mathcal{G}$ and $\boldsymbol\theta_\text{NN}$ are correlated and thus the data from an optimal design for $\boldsymbol{\theta}_\mathcal{G}$ is assumed to be informative in calibrating $\boldsymbol\theta_\text{NN}$. The numerical examples of this work confirm that such an assumption is largely valid, while occasionally the data could be less informative and sometimes even misleading for the calibration of $\boldsymbol\theta_\text{NN}$.

In this work, we propose an ensemble-based indicator that can quantify the data informativeness for the calibration of $\boldsymbol\theta_\text{NN}$. The general concept is to employ an ensemble Kalman method to efficiently approximate the posterior distribution and the utility function involved in Eq.~\eqref{eq: kld utility} and Eq.~\eqref{eq:EIG}, which account for the main computational challenge to standard BED methods when $\boldsymbol{\theta}$ is high-dimensional.

More specifically, ensemble Kalman inversion (EKI) is employed to derive the indicator that quantifies the data informativeness for the calibration of $\boldsymbol{\theta}_\text{NN}$. EKI was developed to solve a Bayesian inverse problem in the general form:
\begin{equation}
    \mathbf{y}=G(\boldsymbol\theta)+\boldsymbol\eta,
\end{equation}
where $G$ denotes a forward map from unknown parameters $\boldsymbol{\theta}$ to the data $\mathbf{y}$, and $\boldsymbol\eta \sim \mathcal{N}(\mathbf{0},\boldsymbol\Gamma)$ denotes the data measurement noises and is often assumed to be with a zero-mean Gaussian distribution. In this work, the forward map $G$ corresponds to a composition of solving the modeled system in Eq.~\eqref{eq:modeled_system} and obtaining the observational data according to a given design $\mathbf{d}$. The parameters $\boldsymbol{\theta}$ only account for the unknown parameters $\boldsymbol\theta_\text{NN}$ of the model discrepancy term, while the parameters $\boldsymbol\theta_\mathcal{G}$ are considered to be fixed.

Given a design $\mathbf{d}$ and the corresponding data measurement $\mathbf{y}$, the EKI updating formula for the ensemble of parameters can be written as:
\begin{equation}
\boldsymbol\theta_{n+1}^{(j)}=\boldsymbol\theta_{n}^{(j)}+\Sigma^{\boldsymbol\theta \mathbf{g}}_n(\Sigma^{\mathbf{g}\mathbf{g}}_n+\Gamma)^{-1}(\mathbf{y}-\mathbf{g}^{(j)}_n),
\end{equation}
where $\mathbf{g}^{(j)}_n:=G(\boldsymbol\theta_{n}^{(j)})$, the index $n$ denotes the $n$-th EKI iteration, and the index $j$ indicates the $j$-th ensemble. The ensemble covariance matrices $\Sigma^{\boldsymbol\theta \mathbf{g}}_n$ and $\Sigma^{\mathbf{g}\mathbf{g}}_n$ can be calculated as:
\begin{equation}
    \begin{aligned}
    &\bar{\boldsymbol\theta}_n = \frac{1}{J}\sum_{j=1}^J \boldsymbol\theta_n^{(j)}, \quad \bar{\mathbf{g}}_n = \frac{1}{J}\sum_{j=1}^J \mathbf{g}^{(j)}_n,  \\
    &\Sigma^{\boldsymbol\theta \mathbf{g}}_n = \frac{1}{J-1} \sum_{j=1}^J \left(\bar{\boldsymbol\theta}_n - \boldsymbol\theta_n^{(j)}\right)\left(\bar{\mathbf{g}}_n  - \mathbf{g}^{(j)}_n \right)^\top ,   \\
    &\Sigma^{\mathbf{g}\mathbf{g}}_n = \frac{1}{J-1} \sum_{j=1}^J \left(\bar{\mathbf{g}}_n  - \mathbf{g}^{(j)}_n \right)\left( \bar{\mathbf{g}}_n  - \mathbf{g}^{(j)}_n \right)^\top.
    \label{eq: eki detials}
\end{aligned}
\end{equation}

With the initial ensemble $\{\boldsymbol\theta_0^{(j)}\}^J_{j=1}$ and the updated ensemble $\{\boldsymbol\theta_K^{(j)}\}^J_{j=1}$ after $K$ times of EKI iterations, we can approximate the KL divergence between the posterior and prior distributions of $\boldsymbol{\theta}$ by treating both ensembles as Gaussian:
\begin{equation}
\begin{gathered}
\label{eq:ensemble_KLD}
        D_{\textrm{KL}}(p(\boldsymbol{\theta}|\mathbf{y}) || p(\boldsymbol{\theta})  ) \approx \tilde{D}_{\textrm{KL}}(\{\boldsymbol\theta_K^{(j)}\}^J_{j=1} || \{\boldsymbol\theta_0^{(j)}\}^J_{j=1}) = \\ 
        \frac{1}{2} \left[ \text{tr} \left( \left( \Sigma^{\boldsymbol{\theta} \boldsymbol{\theta}}_0 \right)^{-1} \Sigma^{\boldsymbol{\theta} \boldsymbol{\theta}}_K \right) - d_{\boldsymbol\theta} + \ln \left( \frac{\det \Sigma^{\boldsymbol{\theta} \boldsymbol{\theta}}_0}{\det \Sigma^{\boldsymbol{\theta} \boldsymbol{\theta}}_K} \right)\right. 
        \left.+ \left(\bar{\boldsymbol{\theta}}_0 - \bar{\boldsymbol{\theta}}_K \right)^\top \left( \Sigma^{\boldsymbol{\theta} \boldsymbol{\theta}}_0 \right)^{-1} \left(\bar{\boldsymbol{\theta}}_0 - \bar{\boldsymbol{\theta}}_K \right) \right],
\end{gathered}
\end{equation}
where $d_{\boldsymbol{\theta}}$ represents the dimension of $\boldsymbol{\theta}$. The ensemble covariance matrices $\Sigma^{\boldsymbol\theta \boldsymbol\theta}_0$ and $\Sigma^{\boldsymbol\theta \boldsymbol\theta}_K$ can be calculated as:
\begin{align}
    &\Sigma^{\boldsymbol\theta \boldsymbol\theta}_n = \frac{1}{J-1} \sum_{j=1}^J \left(\bar{\boldsymbol\theta}_n - \boldsymbol\theta_n^{(j)}\right)\left(\bar{\boldsymbol\theta}_n - \boldsymbol\theta_n^{(j)}\right)^\top.
\end{align}

The ensemble-based approximation of KLD in Eq.~\eqref{eq:ensemble_KLD} can be used to evaluate the informativeness of the data $\mathbf{y}$ from a design $\mathbf{d}$. In this work, we focus on applying the approximated KLD for the high-dimensional parameters $\boldsymbol{\theta}_\text{NN}$ in Eq.~\eqref{eq:modeled_system} and determine whether to perform the calibration of the model discrepancy based on the optimal design and the corresponding measurements from the standard BED of the relatively low-dimensional parameters $\boldsymbol{\theta}_\mathcal{G}$. In practice, a few iterations of EKI updating are sufficient to evaluate the data informativeness, which can effectively avoid performing calibration of the model discrepancies on less informative or even misleading data that could further lead to larger model discrepancies and more biases in the estimation of $\boldsymbol{\theta}_\mathcal{G}$ in Eq.~\eqref{eq:modeled_system}. The selection of these informative design points is iterative: starting from an initial or random candidate, its KLD value and the approximate gradient are determined. New points are then chosen along the KLD ascent direction, and this is repeated until a predefined budget is met. More detailed algorithm to exploit the ensemble-based approximation of KLD to quantify the data informativeness is presented in Algorithm~\ref{alg:ActiveLearning} for the iterative approach introduced in Section~\ref{sec: Model error correction}.

\begin{algorithm}[H]
\caption{Active Learning of Model Discrepancy with BED}
\label{alg:ActiveLearning}
\begin{algorithmic}
    \For {$i=1,2,..., N$} \Comment{Iterating stage}
        \State $\mathbf{d} \gets \argmax_{\mathbf{d} \in \mathcal{D}}\mathbb{E}[U(\boldsymbol{\theta}_{\mathcal{G}},\mathbf{y},\mathbf{d}; \boldsymbol{\theta}_{\textrm{NN}})]$ \Comment{$\boldsymbol{\theta}_{\textrm{NN}}$ is fixed}
        \State $\mathbf{y} \gets \mathbf{u}(\mathbf{d}) + \boldsymbol{\epsilon}$ \Comment{Measurement at optimal design} 
        \State $\boldsymbol\theta_\mathcal{G} \gets \boldsymbol\theta_\mathcal{G}|\mathbf{y}$ \Comment{Bayesian update}
        \State $\boldsymbol\theta_\mathcal{G}^* \gets \argmax_{\boldsymbol\theta_\mathcal{G}} p(\boldsymbol{\theta}_\mathcal{G};\boldsymbol\theta_\text{NN})
        $ \Comment{MAP}
        \State $D_{\textrm{KL}}(p(\boldsymbol{\theta}_{\textrm{NN}}|\mathbf{y}) || p(\boldsymbol{\theta}_{\textrm{NN}}) ) \gets$ \text{EKI} \Comment{Ensemble-based Indicator}
        \If { $D_{\textrm{KL}}$ is large } \Comment{If $\mathbf{y}$ is informative to $\boldsymbol\theta_\text{NN}$}
            \State $\boldsymbol\theta_\text{NN} \gets \argmax_{\boldsymbol\theta_\text{NN}} L(\boldsymbol\theta_\text{NN};\boldsymbol\theta_\mathcal{G}^*,\mathbf{y},\mathbf{d})$ \Comment{$\boldsymbol{\theta}_{\mathcal{G}}^*$ is fixed}
        \EndIf

    \EndFor
    
\end{algorithmic}
\end{algorithm}

\section{Numerical Results}
\label{sec: Numerical Results}

To demonstrate the performance of our proposed method, we study the example of source inversion of a contaminant in the convection-diffusion field, which is a classical test example for BED and has been previously studied in \cite{shen_bayesian_2023}.  
The general goal of this inverse problem is to take concentration measurements in a flow field governed by a convection-diffusion equation and then infer the plume source location. More specifically, the contaminant concentration $\mathbf{u}$ at two-dimensional spatial location $\mathbf{z} = \{z_x, z_y\}$ and time $t$ is governed by the following equation:

\begin{equation}
    \frac{\partial \mathbf{u}(\mathbf{z},t;\boldsymbol\theta)}{\partial t}=D\nabla^2\mathbf{u}-\mathbf{v}(t) \cdot \nabla \mathbf{u}+S(\mathbf{z},t;\boldsymbol\theta),~~~\mathbf{z} \in [z_L,z_R]^2,~~t>0
    \label{eq:true_system_example},
\end{equation}
where $D$ is the diffusion coefficient (assumed to be known with a true value of $1$), $\mathbf{v}=\{v_x,v_y\} \subseteq \mathbb{R}^2$ is a time-dependent convection velocity, $S$ denotes the source term with some parameters $\boldsymbol\theta$. In this work, the true system has an exponentially decay source term in the following form with the parameters $\boldsymbol\theta=\{\theta_x,\theta_y,\theta_h,\theta_s\} \in \mathbb{R}^4$:
\begin{equation}
    S(\mathbf{z},t;\boldsymbol\theta)=\frac{\theta_s}{2\pi\theta_h^2}\exp \left(-\frac{(\theta_x-z_x)^2+(\theta_y-z_y)^2}{2\theta_h^2} \right),
    \label{eq:true_source_term}
\end{equation}
where \(\theta_x\) and \(\theta_y\) denote the source location, \(\theta_h\) and \(\theta_s\) represent the source width and source strength. The initial condition is \(\mathbf{u}(\mathbf{z}, 0;\boldsymbol\theta) = \mathbf{0}\), and a homogeneous Neumann boundary condition is imposed for all sides of the square domain. The convection term is discretized using the second-order total variation diminishing Van Leer scheme, while the diffusion term is treated implicitly via a fast diagonalization method. Time integration follows a semi-implicit scheme: convection is updated explicitly using the forward Euler method, and diffusion is solved implicitly to ensure numerical stability. The time step is chosen as 0.0005 to satisfy the Courant-Friedrichs-Lewy condition imposed by the explicit convection term. The Peclet number in our setup is approximately 0.2, indicating a diffusion-dominated regime where no additional stabilization is required.

Note that in this work, the inversion is performed over a finite-dimensional parameter set $\boldsymbol\theta$ which governs either the source term (e.g., source location, width, and strength) or the parametric form of the convection velocity (e.g., a time-dependent velocity with unknown coefficient $k_y$ in $v_y=k_yt$). The state variable $\mathbf{u}$ not directly inferred but is obtained by solving the governing PDE given $\boldsymbol\theta$ and $k_y$.

\begin{figure}[H]
    \centering
    \includegraphics[width=0.97\linewidth]{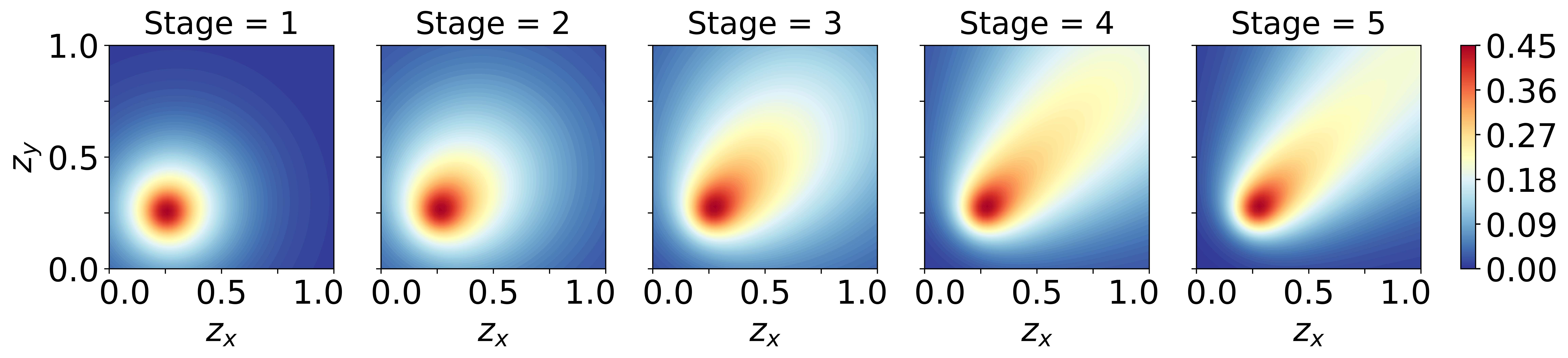}
    \caption{Concentration value at different times in a convection-diffusion field. The numerical simulation is performed in a larger domain ($[-2,3]^2$) and presented in a smaller one ($[0,1]^2$) to emphasize the areas close to the source location. Stages 1-5 correspond to $0.05,0.10,0.15,0.20,0.25$ time units.}
    \label{PDE simulation results}
\end{figure}

\sloppy
For the true system, the parameters in the source term are set as $\theta_s=2$ and $\theta_h=0.05$. Figure~\ref{PDE simulation results} presents the system state $\mathbf{u}$ calculated at $t=0.05,0.10,0.15,0.20,0.25$ time units when the source locates at $\theta_x = 0.25$ and $\theta_y = 0.25$, which illustrates how the source affects the concentration value at different locations across the domain through the convection and diffusion with the time evolution.

The design $\mathbf{d}=\{d_x,d_y,d_t\}$ in this problem refers to the spatiotemporal coordinate to measure the concentration value. More specifically, each design involves the measurement of just one point in the domain. The spatial coordinates of the initial design start at $\{d_x^0,d_y^0\}=(0.5,0.5)$ and will gradually move around to other locations. The differences of spatial coordinates between two consecutive designs (i.e., $d_x^{i+1} - d_x^{i}$ and $d_y^{i+1} - d_y^{i}$) are constrained by the interval $[-0.2, 0.2]^2$. Unlike traditional sensor selection problems where sensors can be placed freely across the entire domain, our setting assumes either a single movable sensor or a localized sensor array that relocates sequentially. The local constraint was initially introduced to ensure that each design decision meaningfully influences the feasible region for future designs, creating a dependency between stages. This local movement constraint is also motivated by a realistic setting where the sensor is movable but has limited relocation capability due to physical or temporal restrictions. The temporal coordinates of the measurement are set with a fixed time step $\Delta t=0.05$ time units. At each time step, the posterior distribution from the previous stage serves as the prior for a new stage of BED as illustrated in Fig.~\ref{fig:graphic_abstract}. The current optimal design also serves as the starting point for the next stage of BED.

In all numerical examples, the physics-based unknown parameters $\boldsymbol{\theta}_\mathcal{G}$ are the location $(\theta_x,\theta_y)$ of the source. The key motivations and conclusions of numerical results are summarized below:
\begin{itemize}
\item We first validate the performance of our approach in an example where the model error only exists as an incorrect value of the parameter $\theta_s$ set in the true source term of Eq.~\eqref{eq:true_source_term}, for which it is feasible to perform the full BED for the joint distribution of unknown parameters $\boldsymbol{\theta}_\mathcal{G}$ and $\theta_s$. The comparison of the estimated model error parameter to the full BED approach confirms the effectiveness of the proposed iterative approach of calibrating the model error. Detailed results can be found in Section~\ref{sec:Correct parametric error}.
\item We study a more challenging example where the model error exists as an incorrect knowledge of the function form for the source term, and such a model discrepancy is characterized by a neural network, which leads to high-dimensional unknown parameters $\boldsymbol{\theta}_\text{NN}$. In practice, the joint distribution of $\boldsymbol{\theta}_\mathcal{G}$ and $\boldsymbol{\theta}_\text{NN}$ is high-dimensional and the full BED would become expensive or even infeasible in such a challenging setup. We demonstrate that the proposed iterative approach provides an efficient and robust correction of model discrepancy and leads to a less biased estimation of $\boldsymbol{\theta}_\mathcal{G}$. Detailed results can be found in Section~\ref{sec:Correct functional error}.

\item We validated the performance of ensemble-based indicator for data informativeness in both of the aforementioned examples. The comparison of indicators computed from different datasets demonstrates that the indicator can effectively identify more informative data for correcting model discrepancy and serves as a much less expensive option to the full Bayesian approach. Detailed results can be found in both Section~\ref{sec:Correct parametric error} and Section~\ref{sec:Correct functional error}.
\item We also examine another more challenging scenario in which the convection coefficient $k_y$ in the velocity field $v_y = k_y t$ is also unknown, an extension beyond the source-term functional-form discrepancy in Section~\ref{sec:Correct functional error}. We compare two strategies: (i) treating $v_y$ as a fixed and incorrect constant, and (ii) jointly calibrating $k_y$ together with the neural network parameters $\boldsymbol{\theta}_\text{NN}$. This example underscores the performance of our framework for an ill-posed problem of unknown parameters estimation. Detailed results can be found in Section~\ref{sec: illposed}.

\end{itemize}

In addition to the numerical example studied in this section, we also test the performance of our method on another numerical example of an acoustic inverse problem and present some key results in~\ref{apd: Numerical case: acoustic inverse}.

\subsection{Validation of Iterative Approach for Parametric Model Error}\label{sec:Correct parametric error}

In this section, the performance of our proposed iterative approach is validated by an example with parametric model error. The setup of this example is designed in such a way that the full BED remains feasible. Compared with the results of full BED approach, we demonstrate that the proposed iterative approach can efficiently and robustly calibrate the parametric model error. More specifically, we set up an example with parametric model error where the true form of the source function in Eq.~\eqref{eq:true_source_term} is known but the value of $\theta_s$ is set incorrectly, e.g., due to the lack of knowledge on the strength of the source term. In addition, the source location $\{\theta_x, \theta_y\}$ is unknown and to be determined by BED in the presence of parametric model error $\theta_s$. The true values are $\theta_x=0.45$, $\theta_y=0.25$, and $\theta_s=2$. The advection velocity is known and set as $v_x=v_y=20t$.

For standard sequential BED, the estimation of those unknown parameters considers a joint probability distribution of $\{\theta_x, \theta_y, \theta_s\}$. For our proposed approach, the joint distribution of $\{\theta_x, \theta_y\}$ is handled by sequential BED, while the estimation of $\theta_s$ is achieved via the gradient-based optimization described in Section~\ref{sec: Model error correction}. It is worth noting that no neural network is used in this example, while the existing model $\mathcal{G}$ depends on $\theta_s$, with which the gradient $\mathrm{d} L/\mathrm{d} \theta_s$ can be derived in a similar way as illustrated in Eq.~\eqref{eq:adjoint_grad}.

\subsubsection{Standard Approach on Parametric Model Error}
\begin{figure}[H]
  \centering
  \includegraphics[width=0.95\linewidth]{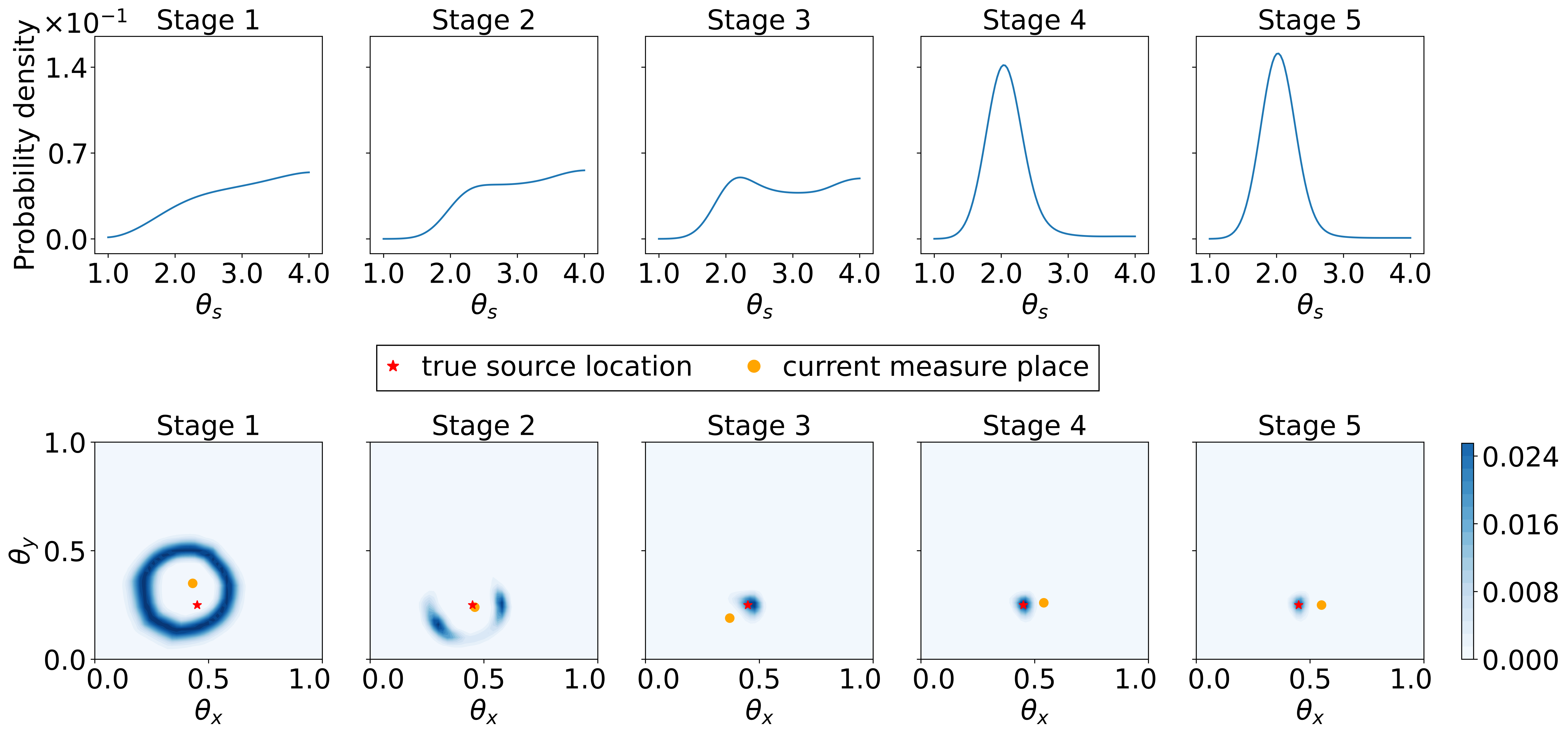}
  \caption{Results of learning parametric model error via standard sequential BED. The top row shows the marginal distribution of $\theta_s$ at each stage, and the bottom row presents the two-dimensional conditional distribution of $\theta_x$ and $\theta_y$, conditioned on the value of $\theta_s$ with the highest marginal probability density.}
  \label{fig:posterior of a six-step 3D OED}
\end{figure}

We first consider joint probability distribution of $\{\theta_x,\theta_y,\theta_s\}$ and perform a five-stage standard sequential BED as the benchmark results. The prior distribution of $\{\theta_x,\theta_y,\theta_s\}$ is assumed to be uniform on the domain $[0,1]\times[0,1]\times[1,4]$, discretized into a $51 \times 51 \times 31$ grid. The five stages correspond to 0.05, 0.10, 0.15, 0.20, and 0.25 time units. For the posterior distribution at each stage, we first present the marginal distribution of $\theta_s$, followed by the two-dimensional conditional distribution of $\theta_x$ and $\theta_y$, conditioned on the value of $\theta_s$ corresponding to the highest marginal probability density. The results in Fig.~\ref{fig:posterior of a six-step 3D OED} demonstrate that the standard approach successfully identifies the true values of the parameters. For $\theta_s$, a peak of probability density gradually forms around the true value of 2 in its marginal distribution, indicating increased confidence in the estimation. In the two-dimensional distribution of $\theta_x$ and $\theta_y$, the high probability region also aligns well with the true source location.

\subsubsection{Hybrid Approach on Parametric Model Error}

In the proposed hybrid approach, we only consider the posterior probability distribution of $\{\theta_x, \theta_y\}$ conditioned on the current stage estimation of model error in the sequential BED, while handling the calibration of model error via a gradient-based optimization introduced in Section~\ref{sec: Model error correction}, thereby restricting the computational cost to a lower-dimensional probability distribution. 

At the first stage of sequential BED, we employ a uniform prior distribution of $\{\theta_x, \theta_y\}$ over the range $[0,1]^2$, discretized into a $51\times 51$ grid of points, and set an initial value of $\theta_s = 3$. In each stage, BED is performed to infer the posterior distribution of $\{\theta_x, \theta_y\}$, and the optimal design is determined accordingly based on the information gain introduced in Section~\ref{sec:OED}. The optimal design subsequently provides additional information that facilitates the calibration of the model error $\theta_s$. In terms of the gradient-based optimization of model error, we have tested both automatic differentiation and adjoint method to acquire the gradient of $\mathrm{d} L/\mathrm{d} \theta_s$ and achieved comparable results. For simplicity and clarity, we only show the results from the automatic differentiation approach, where we have developed a numerical solver based on JAX-CFD~\cite{jax2018github, Kochkov2021-ML-CFD} to facilitate the extraction of gradient information.

\begin{figure}[H]
  \centering
  \includegraphics[width=0.95\linewidth]{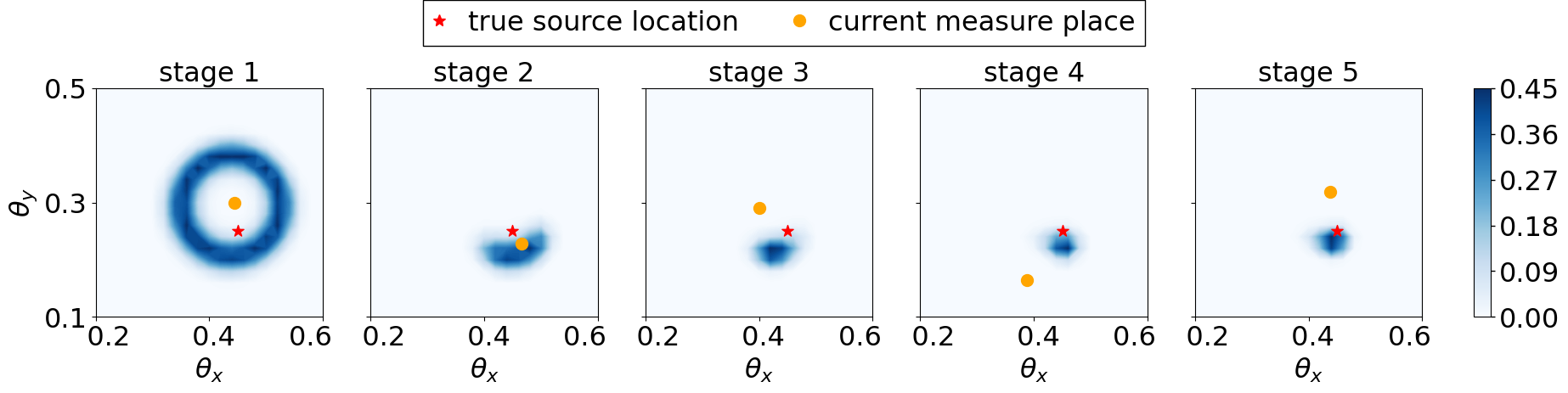}
  \caption{Posterior results of BED parameter $\{\theta_x,\theta_y\}$ via the hybrid approach on parametric error. The $\{\theta_x,\theta_y\}$ space is $[0,1]^2$ as before, with a zoomed-in view $[0.2,0.6]\times[0.1,0.5]$ to highlight detailed behaviors.}
  \label{fig: posterior of learning parameter error}
\end{figure}

Figure~\ref{fig: posterior of learning parameter error} shows the evolution of the posterior distribution for $\{\theta_x, \theta_y\}$ across five stages. The high probability regions gradually converge to the true source location, suggesting an effective correction of parametric model error $\theta_s$. The comparison between Figs~\ref{fig:posterior of a six-step 3D OED} and \ref{fig: posterior of learning parameter error} confirms that the proposed hybrid approach achieves similar results to standard BED for the joint distribution of $\{\theta_x, \theta_y,\theta_s\}$ but with lower computational cost. Specifically, in the standard full approach, the posterior is approximated over a discretized grid of size $51\times51\times31$, requiring 80,631 forward model evaluations per stage. In contrast, the proposed hybrid approach only requires $51\times51=2601$ evaluations per stage, leaving the task of parameter estimation for $\theta_s$ to the gradient-based optimization as described in Section~\ref{sec: Model error correction}. It is worth noting that the proposed hybrid approach has more computational advantage than the full Bayesian approach when structural model error exists, for which the model error needs to be parametrized by higher-dimensional unknown parameters and we provide a comprehensive study in Section~\ref{sec:Correct functional error}.

\begin{figure}[H]
  \centering
  \begin{subfigure}[b]{0.4\textwidth}
    \centering
    \includegraphics[width=\linewidth]{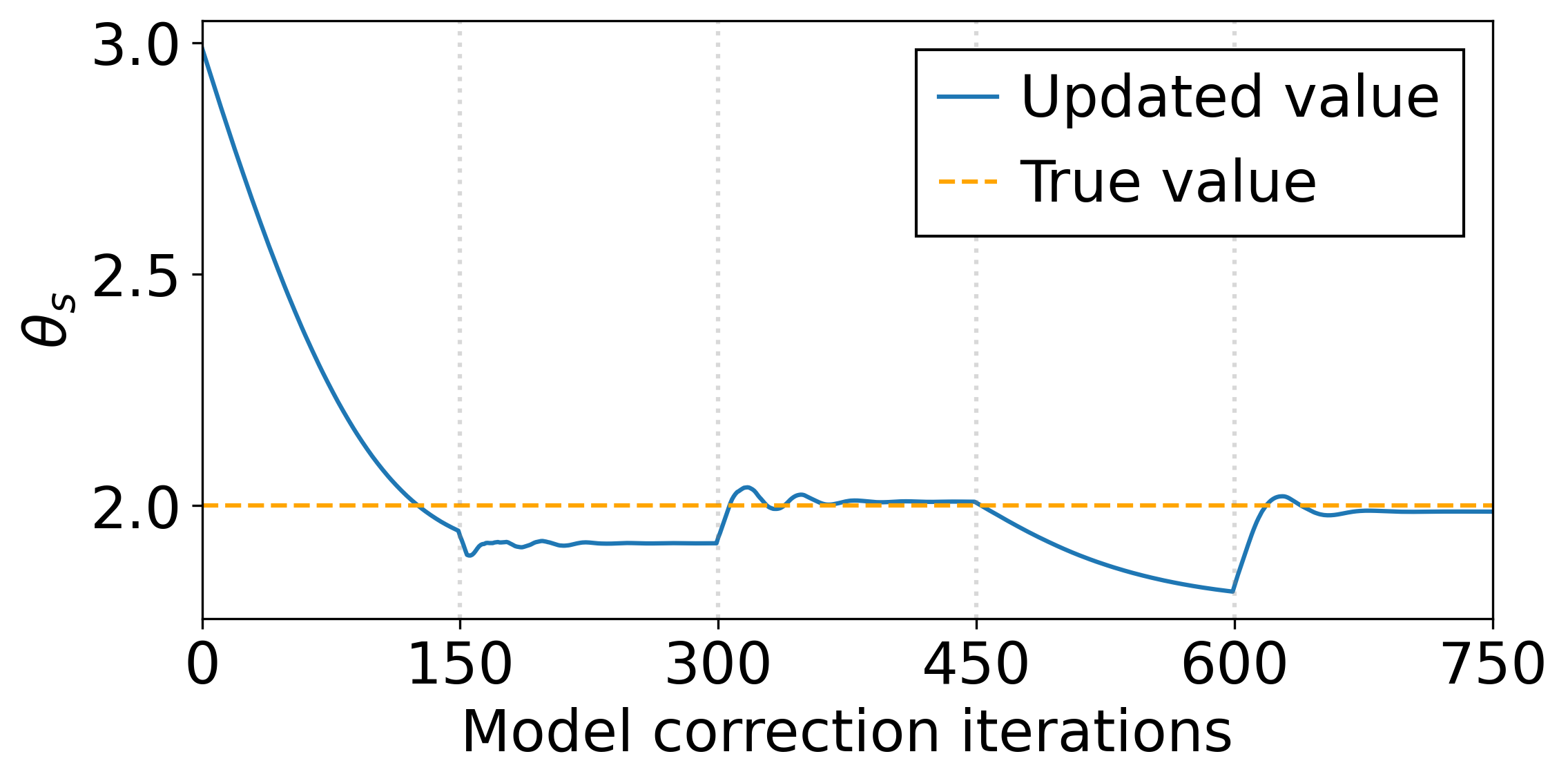}
    \caption{Original model correction}
  \label{Results of learning parameter value}
  \end{subfigure}%
  \begin{subfigure}[b]{0.6\textwidth}
    \centering
    \includegraphics[width=0.95\linewidth]{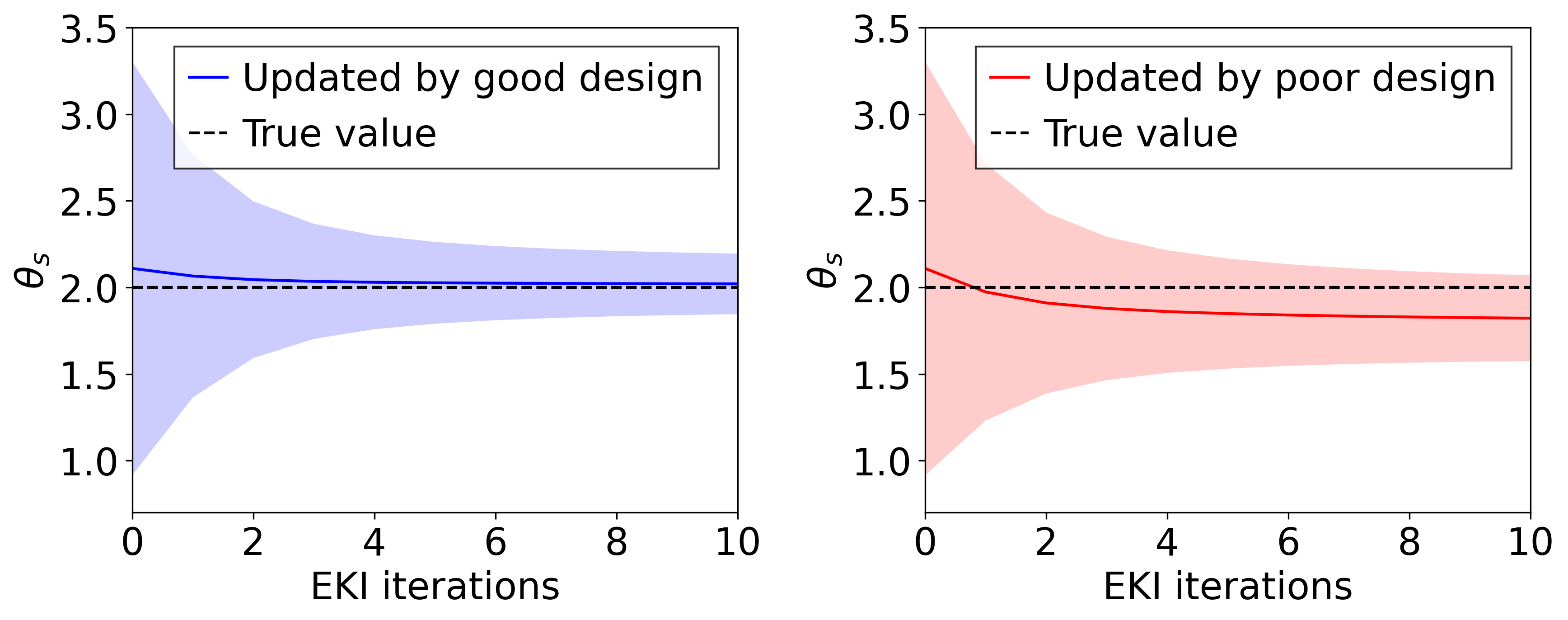}
    \caption{EKI updating}
  \label{Results of EKI in learning parameter error}
  \end{subfigure}
  \caption{Results of parametric model error calibration via the hybrid approach and the ensemble-based indicator of data informativeness. Panel (a) represents the model error calibration process at five stages, with each stage having 150 iterations of gradient-based optimization. Panel (b) presents the ensemble-based error parameter results updated by two different designs at stage 4.}
  \label{parameter Results of learning parameter error}
\end{figure}

We further investigate the results of corrected parametric model error $\theta_s$. As shown in Fig.~\ref{Results of learning parameter value}, the proposed hybrid approach identifies the true value of $\theta^\dagger_s = 2$. However, the estimate of $\theta_s$ tends to converge towards an incorrect value between iterations 450 and 600 (stage 4) during the optimization process. This deviation highlights the impact of experimental design quality on model error correction. More specifically, different designs yield varying data informativeness for $\theta_s$: some lead to robust estimation toward the true value, while other introduces bias. It is worth noting that the data quality cannot be quantified with the amount of biases in either the corrected model $\mathcal{G}+\mathrm{NN}$ or the simulated system states, as the truth of both is unknown \textit{a priori}. To address this issue, we employ the approximated KLD defined in Eq.~\eqref{eq:ensemble_KLD} to practically evaluate the data informativeness. Specifically, the original design location at stage 4 is $\{0.44,0.32\}$, which we refer to as the poor design due to its limited ability to reduce the bias in $\theta_s$. On the other hand, we randomly check several nearby points and find that better calibration results of $\theta_s$ can be achieved by a nearby alternative design point $\{0.41,0.26\}$, which is considered to yield more informative data and is referred to as the good design. In Fig.~\ref{Results of EKI in learning parameter error}, we illustrate how $\theta_s$ evolves under the two different designs by plotting the ensemble mean and variance at each EKI iteration. The poor-quality design at stage 4 not only renders a bias in the corrected value for the parametric model error $\theta_s$ but also leads to a slower rate of variance reduction during EKI iterations, which implies a smaller approximated KLD between the prior and posterior ensemble of $\theta_s$. The comparison in Fig.~\ref{Results of EKI in learning parameter error} confirms that the performance of the ensemble-based approximated KLD as an indicator to evaluate data informativeness for the calibration of parametric model error.

\subsection{Hybrid Approach on Structural Errors}
\label{sec:Correct functional error}

In this section, we focus on the scenario that structural error exists in the modeled system of Eq.~\eqref{eq:modeled_system}, i.e., $\mathcal{G}^\dagger - \mathcal{G}$ is non-negligible. The structural model error is characterized by a neural-network-based model as shown in Eq.~\eqref{eq:modeled_system} with parameters $\boldsymbol{\theta}_\text{NN}$ in a high-dimensional space. This scenario presents a challenge for the BED methods, which often become prohibitively expensive or even infeasible due to the computational cost of handling high-dimensional parameter spaces. Our goal here is to demonstrate the efficacy of the proposed method in providing an efficient and robust correction of structural error based on the optimal designs and the corresponding data from BED for the relatively low-dimensional parameters $\boldsymbol{\theta}_\mathcal{G}$.

The modeled system is in the same form of Eq.~\eqref{eq:true_system_example}, with the source term defined by:
\begin{equation}
    S(\mathbf{z},t;\bm\theta)=\frac{3\theta_s}{\pi\left(\frac{(\theta_x-z_x)^2+(\theta_y-z_y)^2}{2\theta_h^2}+2\theta_h^2\right)},
    \label{eq: c sourse}
\end{equation}
which is different from the exponentially decaying source term as defined in Eq.~\eqref{eq:true_source_term} for the true system. Without addressing this model structural error, the inference of source location through standard BED yields biased results. In this work, a neural network $\text{NN}(z_x-\theta_x,z_y-\theta_y;\bm{\theta}_\text{NN})$ is employed to characterize the model structural error. More specifically, we utilize a fully connected neural network consisting of 37 parameters, a configuration that introduces a higher-dimensional parameter space than the example of parametric model error in Section~\ref{sec:Correct parametric error}. This network takes a 2-dimensional input vector, representing the relative distance between a given grid point and the source location, to output a scalar correction at each spatial point. This output quantifies the difference between the true and modeled source term values. The architecture features two hidden layers, each with four neurons and a tanh activation function. Since the governing equation's RHS source term follows an inherently pointwise exponential or reciprocal decay function, a fully connected feedforward neural network mapping individual points independently is a natural choice. Since our coordinate‐based network is trained only on a limited set of pointwise data from the BED framework and optimized without any gradient‐ or residual‐based loss terms, it cannot reliably capture the non-local spatial information for characterizing convection and diffusion. On the other hand, the neural network output is time-invariant, while the convective velocity increases over time, it is incapable of matching the full RHS alone and thus cannot replace the role of existing physics-based terms in the modeled system. In addition, a structural constraint is imposed such that the neural network approximates a spatially decaying function that centers at its input reference point, i.e., the estimated source location $\{\theta_x,\theta_y\}$, which is the same center as the physical-based source term in the RHS of the governing equation. In our numerical example, this constraint is incorporated via a pre-training phase, i.e., the network is initialized on data generated from the true target decaying source form. It is worth noting that such pre-training data would not guarantee the desired constraint in the subsequently trained model and may not be available in real-world applications. An alternative approach to enforce desired constraints on the trained neural network model is to add a penalty term into the loss function, e.g., penalizing the positive gradient with respect to the outward direction from the reference center point in this numerical example, to ensure the trained neural network is a spatially decaying function.

We employ a uniform prior distribution over the range $[0,1]^2$ for $\{\theta_x, \theta_y\}$, discretized into a $51\times 51$ grid, with the true value of the source location set at $\{0.25,0.25\}$. The values of $\{\theta_s^\dagger, \theta_h^\dagger\}$ are set as $\{2, 0.05\}$ in both the true system and the modeled one. The advection velocity is known and set as $v_x=v_y=50t$. Design constraint remains the $[-0.2,0.2]^2$.

\begin{figure}[H]
  \centering
  \begin{subfigure}[b]{1\textwidth}
    \centering
    \includegraphics[width=0.95\linewidth]{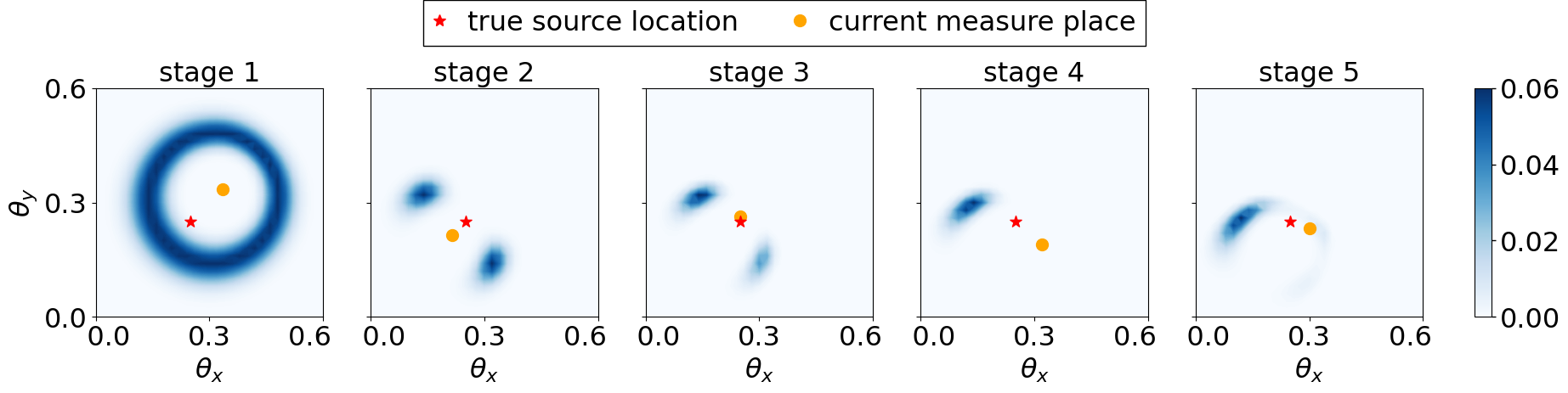}
    \caption{With $\boldsymbol\theta_\text{NN}$ frozen}
    \label{fig:posterior_DP_1}
  \end{subfigure}%
  \vspace{10pt}
  \begin{subfigure}[b]{1\textwidth}
    \centering
    \includegraphics[width=0.95\linewidth]{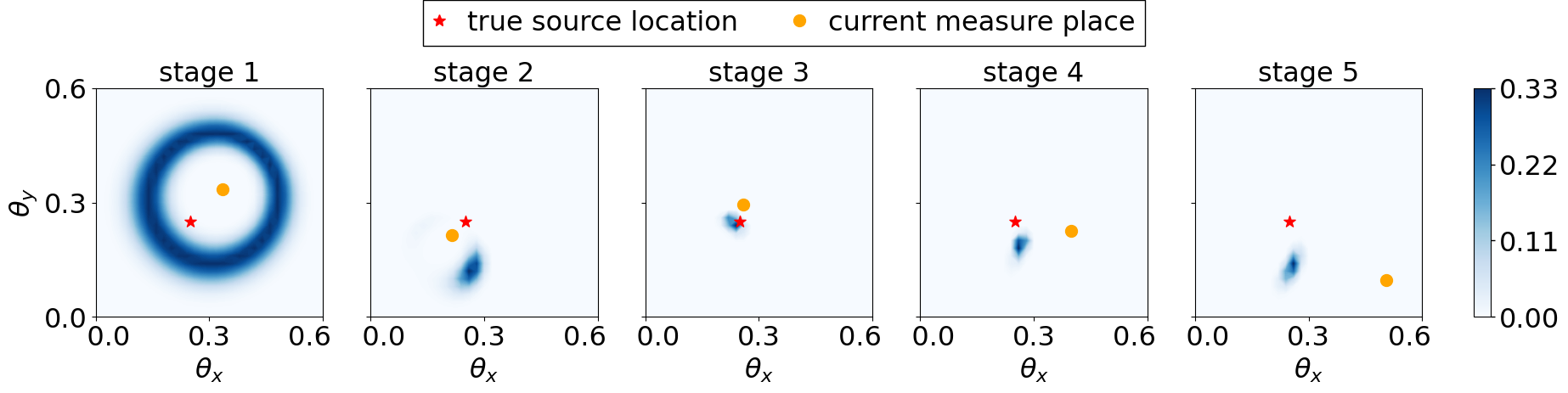}
    \caption{Update $\boldsymbol\theta_\text{NN}$}
    \label{fig:posterior_DP_2}
  \end{subfigure}%
  \caption{Posterior distributions of inferred source location $\{\theta_x,\theta_y\}$ based on (a) a randomly initialized structural error term and (b) the calibrated structural error term.}
  \label{fig: correct network error}
\end{figure}

We first present the results of standard sequential BED for physics-based parameters updated without correcting the structural error. It is expected that model discrepancy introduces bias in both the model forecast and likelihood values, resulting in an incorrect posterior distribution, i.e., the high probability area does not cover the true source location in Fig.~\ref{fig:posterior_DP_1}. Fig.~\ref{fig:posterior_DP_2} shows that the posterior distribution can be improved by applying the proposed hybrid framework to correct the model error, and the high probability areas gradually converge to the true source location from stage 1 to stage 3. The deviation of high probability areas from the true source location in stages 4 and 5 is mainly caused by the uninformative design and data used for model error correction at stage 3. These design and data, optimized for physical parameters at stage 3, introduce biases into the calibrated model, which propagate into the Bayesian updating of physical parameters in subsequent stages. This observation does not contradict the Bayesian framework, as the posterior may still shift away from the prior if the likelihood is sharply concentrated in regions with non-negligible prior mass.

\begin{figure}[H]
  \centering
  \begin{subfigure}[b]{1\textwidth}
    \centering
    \includegraphics[width=0.95\linewidth]{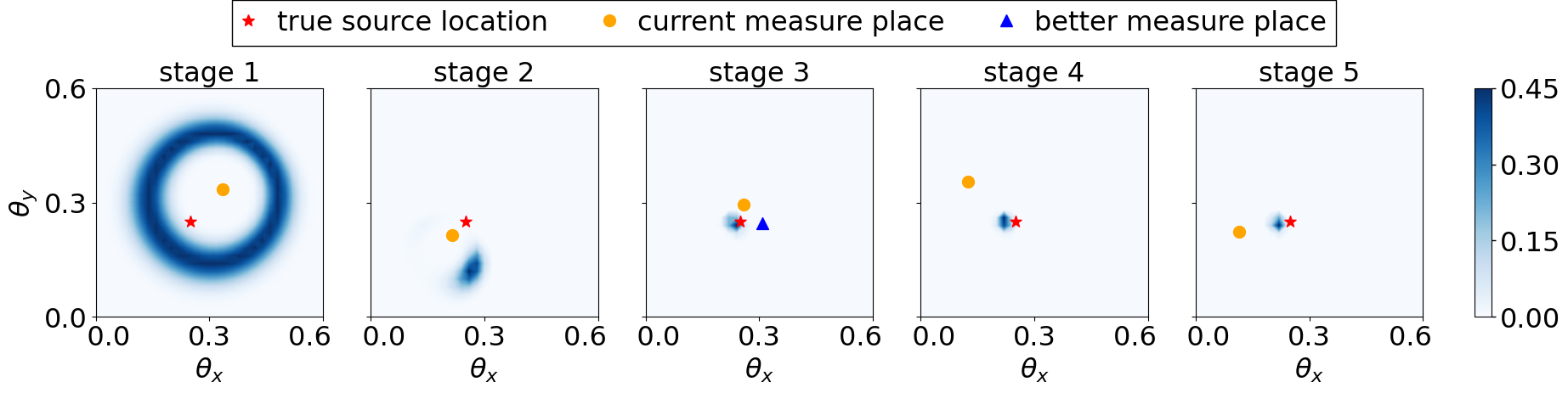}
  \end{subfigure}%
  \caption{Posterior distributions of inferred source location $\{\theta_x,\theta_y\}$ based on the calibrated structural error using more informative data at stage 3. The blue triangular indicates the better design for structural error correction.}
  \label{fig: enhanced correct network error}
\end{figure}

\begin{figure}[H]
  \centering
  \begin{subfigure}[b]{0.6\textwidth}
    \centering
    \includegraphics[width=0.95\linewidth]{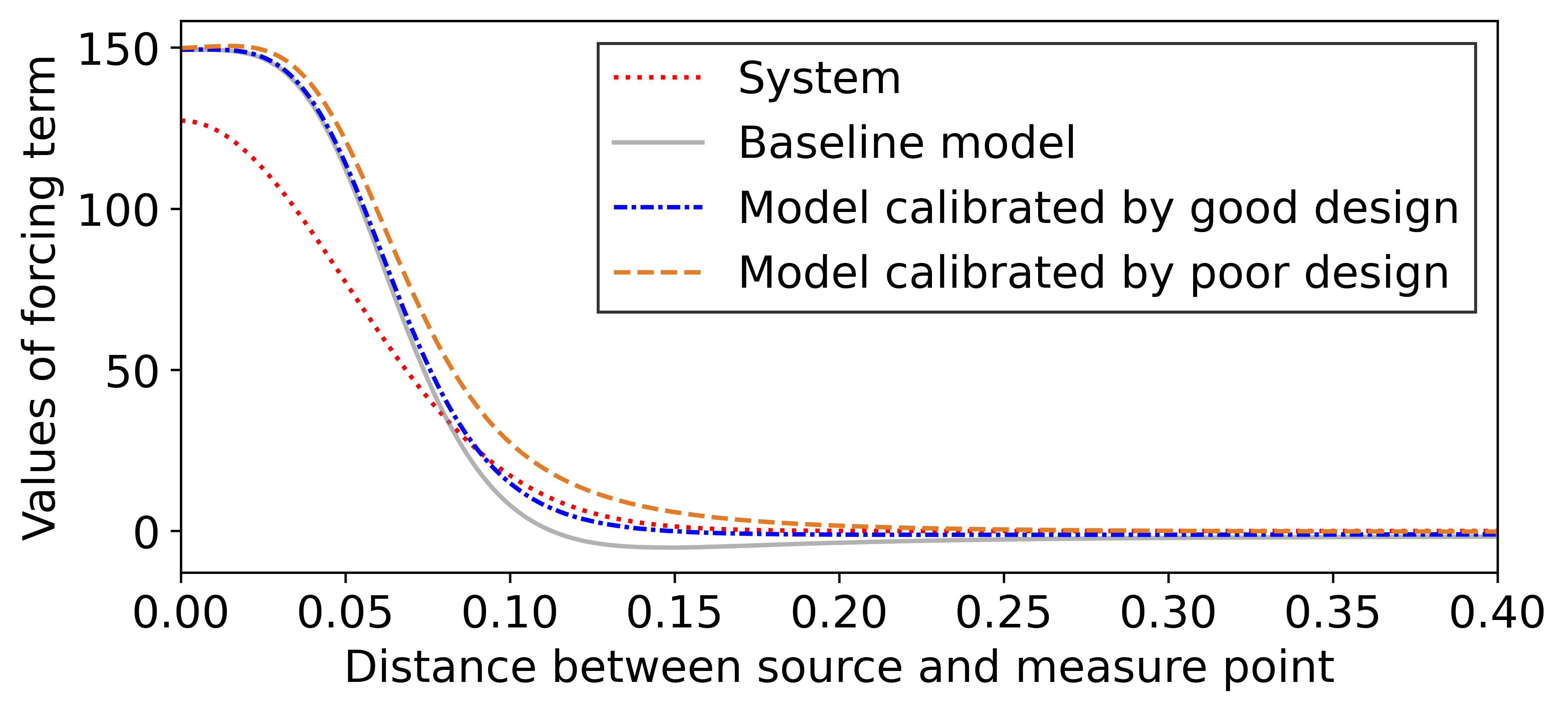}
    \caption{Network performance}
  \label{Results of calibrating neural network}
  \end{subfigure}%
  \begin{subfigure}[b]{0.4\textwidth}
    \centering
    \includegraphics[width=0.76\linewidth]{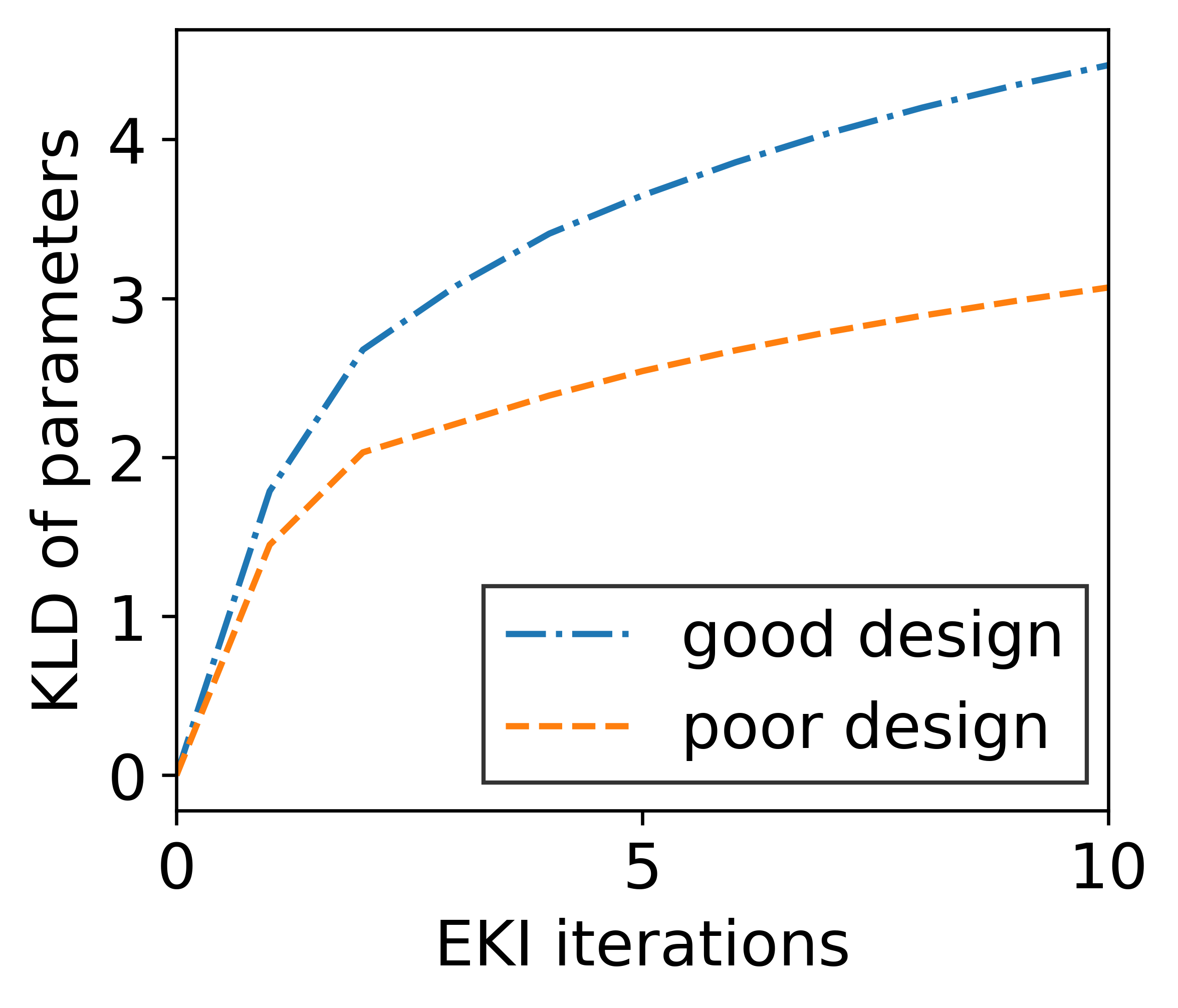}
    \caption{EKI results}
  \label{Results of EKI in learning structural error}
  \end{subfigure}
  \caption{Results of calibrating the structural error via the hybrid approach. Panel (a) presents the comparison of the true source term and the modeled ones. Panel (b) presents the ensemble-based approximated KLD for the model error calibration with good and poor designs.}
  \label{parameter Results of learning structural error}
\end{figure}

Figure~\ref{fig: enhanced correct network error} shows that a more informative design and the corresponding measurement data (adopted at stage 3) for the model error correction significantly improve the posterior distributions of inferred source location at stages 4 and 5. We present the results using the ensemble-based approximated KLD to assess the data informativeness of a given design in Fig.~\ref{parameter Results of learning structural error}. Two designs are selected for comparison: (i) the design identified by BED for posterior estimation at stage 3, represented by the orange dot in Fig.~\ref{fig: enhanced correct network error}, and (ii) a modified design represented by the blue triangle in the same panel. With the untrained model at stage 3 as a baseline, we first generate an ensemble of its parameters and apply EKI to update the ensemble for 10 iterations using the two designs and corresponding measurement data. The ensemble-based approximated KLD between the updated and initial ensembles at each EKI iteration in Fig.~\ref{Results of EKI in learning structural error} shows a faster increase with the good design, which confirms that the ensemble-based indicator proposed in Section~\ref{sec:ensemble_indicator} is effective of assessing data informativeness. The model error updated by the good design also shows less deviation from the true discrepancy in Fig.~\ref{Results of calibrating neural network}. Considering that the measurement data at a few design points are not sufficient to correct the model error term across the whole domain, some remaining differences in Fig.~\ref{Results of calibrating neural network} between the true source term and the modeled one are expected. Even with those differences in the modeled source term, especially at a smaller distance from the source location, the posterior distributions in Fig.~\ref{fig: enhanced correct network error} still demonstrate a good estimation of the inferred source location.

With the structural model errors calibrated by the good and poor designs, we further investigate the performance of the solution field. More specifically, we solve for the modeled systems and compare the solutions with the true solution field. To quantitatively assess the performance of the solution fields, we also employ the mean squared error (MSE) and relative error (RE) of a model solution field $\mathbf{u}$ against the true solution field $\mathbf{u}^\dagger$:
\begin{equation}
    \begin{aligned}
        \text{MSE}&=\frac{1}{N_z}\sum_{z_x,z_y} (\mathbf{u}(z_x,z_y) - \mathbf{u}^\dagger(z_x,z_y))^2,\\
        \text{RE}&=\frac{\sum_{z_x,z_y} |\mathbf{u}(z_x,z_y) - \mathbf{u}^\dagger(z_x,z_y)|}{\sum_{z_x,z_y}  |\mathbf{u}^\dagger(z_x,z_y)| },
    \end{aligned}
    \label{eq:define error}
\end{equation}
where the $N_z$ represents the total number of discretization points in the solution field.

\begin{figure}[H]
  \centering
  \begin{subfigure}[b]{1\textwidth}
    \centering
    \includegraphics[width=0.9\linewidth]{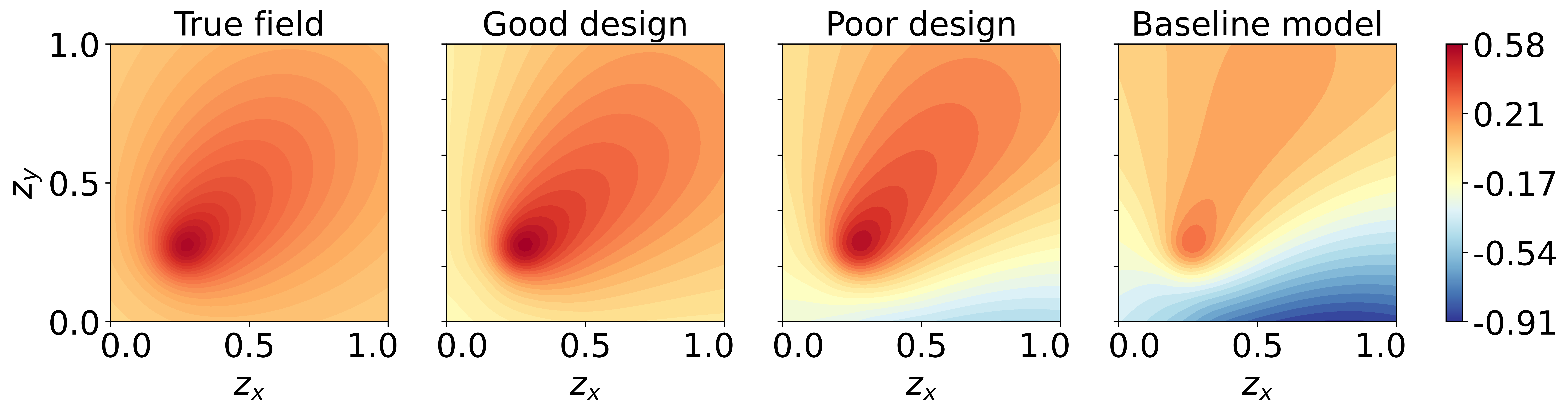}
    \caption{Predicted fields}
  \end{subfigure}%
  \vspace{10pt}
  \begin{subfigure}[b]{1\textwidth}
    \centering
    \includegraphics[width=0.7\linewidth]{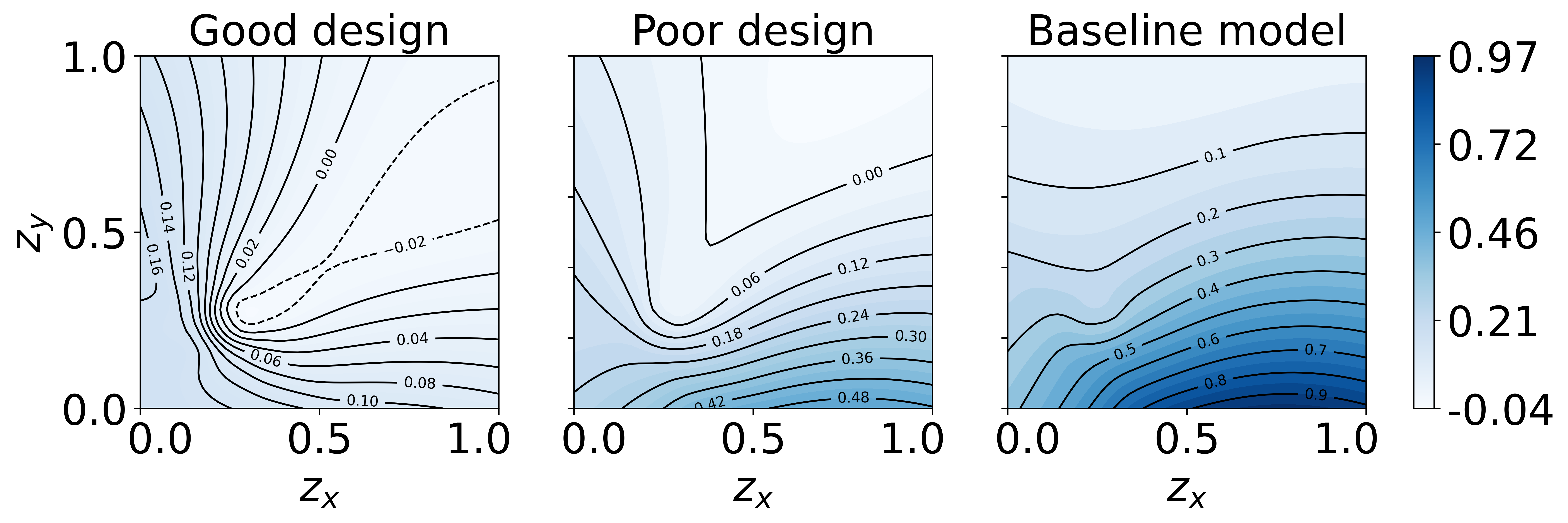}
    \caption{Deviation between the predicted and true fields}
  \end{subfigure}%
  \caption{Solution fields of the true system and the modeled ones. Panel (a) shows a visual comparison of the solutions fields. Panel (b) presents the mismatches between the true solution field and the solution fields from three different modeled systems.}
  \label{fig: predict field}
\end{figure}

Figure~\ref{fig: predict field} and Table~\ref{tab: quantitative} present the comparison between the solutions from the modeled systems and the true one. Both the good design and the poor one lead to improvements in the agreement with the true solution field over the baseline model. However, the good design achieves a better agreement with the true field, as evidenced by the lower deviation in the contour of Fig.~\ref{fig: predict field} and the smaller MSE and RE values in Table~\ref{tab: quantitative}. This improvement confirms a more effective correction of model discrepancy, which also explains the improved performance of inferred source location in Fig.~\ref{fig: enhanced correct network error}.

\begin{table}[H]
    \centering
    \caption{Mean squared error and relative error of the solution fields based on the modeled systems.}
    \begin{tabular}{|c|c|c|c|}
        \hline
        & Good design & Poor design & Baseline model \\ \hline
        MSE & 0.005 & 0.032 & 0.142 \\ \hline
        RE & 0.297 & 0.665 & 1.553 \\ \hline
    \end{tabular}
    \label{tab: quantitative}
\end{table}

\subsection{Hybrid Approach on Ill‑Posed Problems}\label{sec: illposed}

In this section, we present a more challenging numerical example that aims at illustrating the performance of our method in an ill‑posed setting, i.e., the solution of the physics-based parameters and the neural-network-based structural error is not unique. 
Extending from the structural error case presented in Section~\ref{sec:Correct functional error}, we further assume incorrect prior knowledge in the linearly increasing convection velocity $v(t)$. Specifically, the accurate velocity on $x$-axis is known, whereas the growth rate $k_y$ in $v_y = k_y t$ is known as a wrong value. Intuitively, a stronger convective component along one axis deflects the passive scalar plume toward the orthogonal direction, producing an asymmetrically skewed trajectory in the scalar field. This setting increases the nonlinearity of the inverse problem, since the convection coefficient \(k_y\) enters the advective operator \(v_y\partial_y = (k_y t)\partial_y\), making the map from \(k_y\) to the observed field values inherently nonlinear rather than a simple scalar scaling of the data.

The source location $\{\theta_x, \theta_y\}$ remains unknown, and the previously assumed incorrect source form is retained under network correction $\boldsymbol{\theta}_\text{NN}$. Following the workflow in Section \ref{sec:Correct functional error}, this problem can be solved by treating $\{\theta_x, \theta_y, k_y\}$ as physical parameters and introducing a network with $\boldsymbol{\theta}_\text{NN}$ to capture the structural error. To study an ill-posed scenario, only the parameters \(\{\theta_x, \theta_y\}\) are modeled as random variables characterized by probability distributions and are utilized for identifying optimal design, while the parameters \(\{k_y, \boldsymbol{\theta}_\text{NN}\}\) are updated jointly via gradient-based methods. The ill-posedness arises from the flexibility of the neural network, i.e., the network is sufficiently expressive to match the observed data even when \(k_y\) takes an incorrect value. Consequently, many pairs of \(\{k_y,\boldsymbol{ \theta}_\text{NN}\}\) can yield simulation results that match the observed data. It is worth noting that the ill-posedness discussed here is different from that in Section \ref{sec:Correct functional error}, i.e., the trained neural network can compensate or even completely replace the role of existing physics-based terms in the modeled system, thus making $\theta_\mathcal{G}$ essentially non-identifiable. In this section, we focus on studying the possible ill-posedness in the optimization stage, which is an internal non-identifiability that emerges during the joint optimization of certain physical parameters (e.g., $k_y$) with the network parameters ($\theta_\text{NN}$). This challenging scenario is also significantly exacerbated by the limited amount of training data at every optimization stage. We first demonstrate the ill-posedness experimentally by fixing $k_y$ at an incorrect value and only updating $\boldsymbol{\theta}_\text{NN}$. Subsequently, we illustrate that the inability to precisely identify the true value of $k_y$ does not harm the inference of $\{\theta_x, \theta_y\}$.

For the detailed setup, the true source parameters $\{\theta_x, \theta_y\}$ is set as $\{0.25, 0.35\}$ with a uniform prior distribution on $[0,1]^2$. The true value of the growth rate $k_y$ is $50$ while the initially assumed incorrect value is $70$. The structural error in the source term, the neural network architecture, and other conditions remain consistent with those in Section~\ref{sec:Correct functional error}. Instead of employing a single measurement location, we adopt a centered $3\times3$ sensor grid consisting of the original point and its eight immediate neighbors at intervals of 0.06. This increases the total amount of information in the data, thus alleviating the intrinsic challenge of jointly learning $k_y$ and the structural error. This setup retains physical interpretability, as sensor arrays in small regions or a rapidly movable sensor (with measurement times shorter than system evolution times) are both common in the experimental design community.

\begin{figure}
  \centering
  \begin{subfigure}[b]{1\textwidth}
    \centering
    \includegraphics[width=0.95\linewidth]{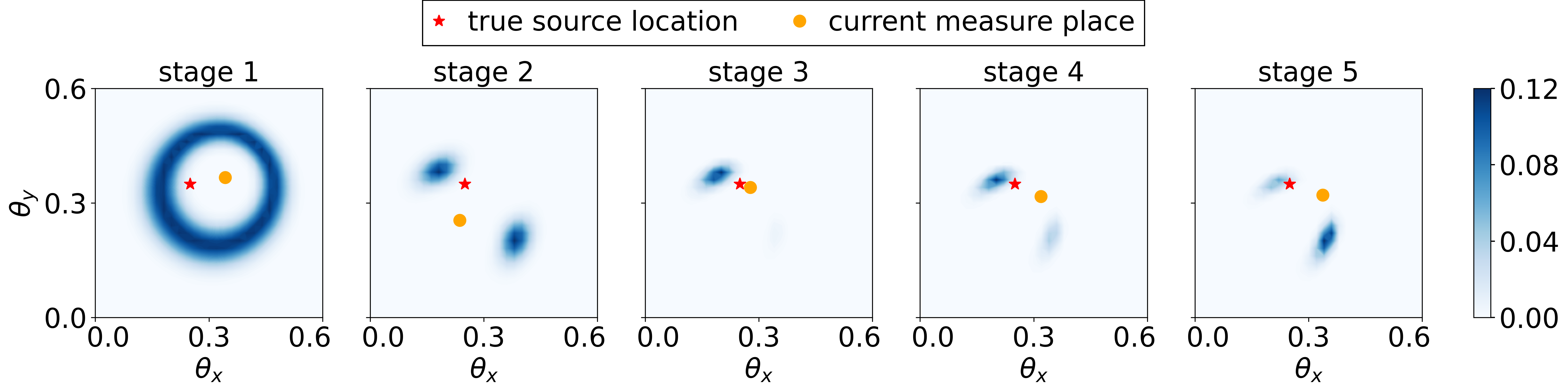}
    \caption{With $k_y$ and $\theta_\text{NN}$ frozen}
    \label{fig:posterior_no_correction}
  \end{subfigure}%
  \vspace{10pt}
  \begin{subfigure}[b]{1\textwidth}
    \centering
    \includegraphics[width=0.95\linewidth]{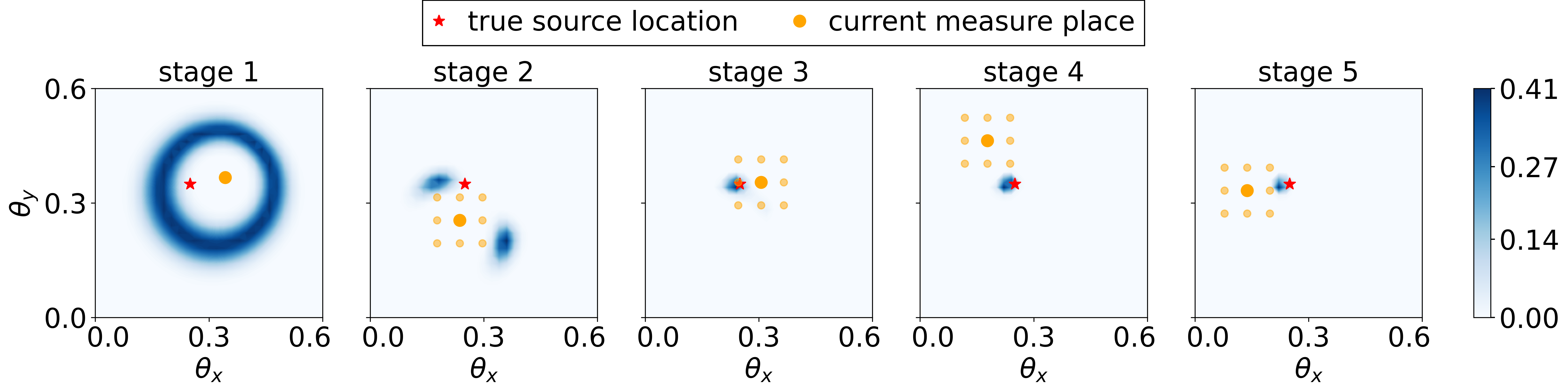}
    \caption{With $k_y$ frozen}
    \label{fig:posterior_fix_vel}
  \end{subfigure}%
  \vspace{10pt}
  \begin{subfigure}[b]{1\textwidth}
    \centering
    \includegraphics[width=0.95\linewidth]{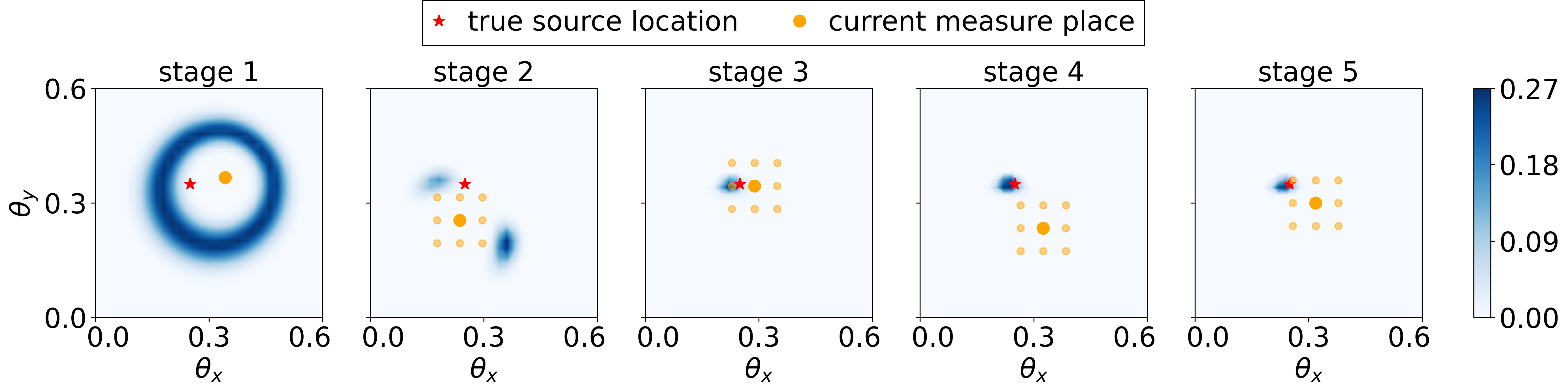}
    \caption{Update $k_y$}
    \label{fig:posterior_not_fix}
  \end{subfigure}%
  \caption{Posterior distributions of inferred source location $\{\theta_x,\theta_y\}$ with (a) both $k_y$ and $\theta_\text{NN}$ fixed, (b) $k_y$ fixed and $\boldsymbol{\theta}_\text{NN}$ updated , and (c) $k_y$ and $\theta_\text{NN}$ jointly updated. In panel (a), no model correction is applied; only one design point is selected per stage using BED, resulting in a single observation per plot. In panel (b) and (c), larger orange dots indicate design points used for BED, while smaller semi-transparent orange dots denote additional observations used for model correction. }
  \label{fig: correct network error 2}
\end{figure}

We first consider the case in which $k_y$ is fixed at an incorrect value and $\theta_\text{NN}$ is updated. Figure~\ref{fig: correct network error 2} presents the inference results of $\{\theta_x, \theta_y\}$ using different models. Without the correction of $\theta_\text{NN}$, as shown in Figure~\ref{fig:posterior_no_correction}, the model with biased convection speed and incorrect source form yields a biased and multi-modal posterior distribution. However, with an updating network $\boldsymbol{\theta}_\text{NN}$ and fixed biased $k_y$, the highlighted areas in Figure~\ref{fig:posterior_fix_vel} progressively converge toward the true source location, demonstrating that our framework accurately identifies the source position. We attribute this improvement to the neural network for compensating the incorrect convection term as well. To verify this argument, we compute the relative errors of the simulated solution fields and the PDE right-hand side (RHS) as defined in Eq.~\eqref{eq:define error}, excluding the diffusion term, at various update stages before and after correction. The results with fixed $k_y$ in Table~\ref{tab: nn case 2} show that the data-driven correction consistently reduces the RHS error, even with an incorrect $k_y$ value, thereby enhancing the accuracy of the simulated fields in the vicinity of the data. Optimizing $k_y$ allows for directly adjusting the incorrect physical convection characteristics of the modeled system, leading to more significant error reduction. It is expected that the neural network designed to model a spatially decaying source correction term is inherently limited in its ability to globally compensate for an inaccurate convection velocity, compared to the direct adjustment of $k_y$.

\begin{table}[H]
    \centering
    \caption{Relative errors of the simulated solution fields and the PDE right-hand-side (RHS) without diffusion at different update stages. “No” refers to the model before applying data-driven correction at the current stage, while “Yes” refers to the corrected model after incorporating data at the same stage. The RHS error corresponds to the right-hand side of the PDE, excluding the diffusion term, as the diffusion term is same between model and system. All errors are computed over a small region slightly larger than the observation area.}
    \label{tab: nn case 2}
    \resizebox{\textwidth}{!}{
    \begin{tabular}{|c|c|c|c|c|c|c|c|c|}
    \hline
    Stage Index & \multicolumn{2}{|c|}{2} & \multicolumn{2}{|c|}{3} & \multicolumn{2}{|c|}{4} & \multicolumn{2}{|c|}{5} \\ \hline
    Model Correction & No & Yes & No & Yes & No & Yes & No & Yes \\ \hline
    \multicolumn{9}{|c|}{Fix $k_y$}\\ \hline
    RE of simulated field       & 0.4951& 0.1544& 0.3575& 0.0687& 0.2552& 0.1534& 0.7752& 0.2742\\ \hline
    RE of RHS*       & 0.6378& 0.6248& 0.5069& 0.4514& 0.6224& 0.5895& 0.6971& 0.6968\\ \hline
    \multicolumn{9}{|c|}{Update $k_y$}\\ \hline
    RE of simulated field       & 0.4951& 0.1003& 0.2343& 0.1069& 0.7091& 0.1037& 0.2461& 0.1159\\ \hline
    RE of RHS*        & 0.6378& 0.5664& 0.6479& 0.6122& 0.4831& 0.3555& 0.4225& 0.4413\\ \hline
    \end{tabular}
    }
\end{table}

\begin{figure}
  \centering
  \begin{subfigure}[b]{1\textwidth}
    \centering
    \includegraphics[width=0.95\linewidth]{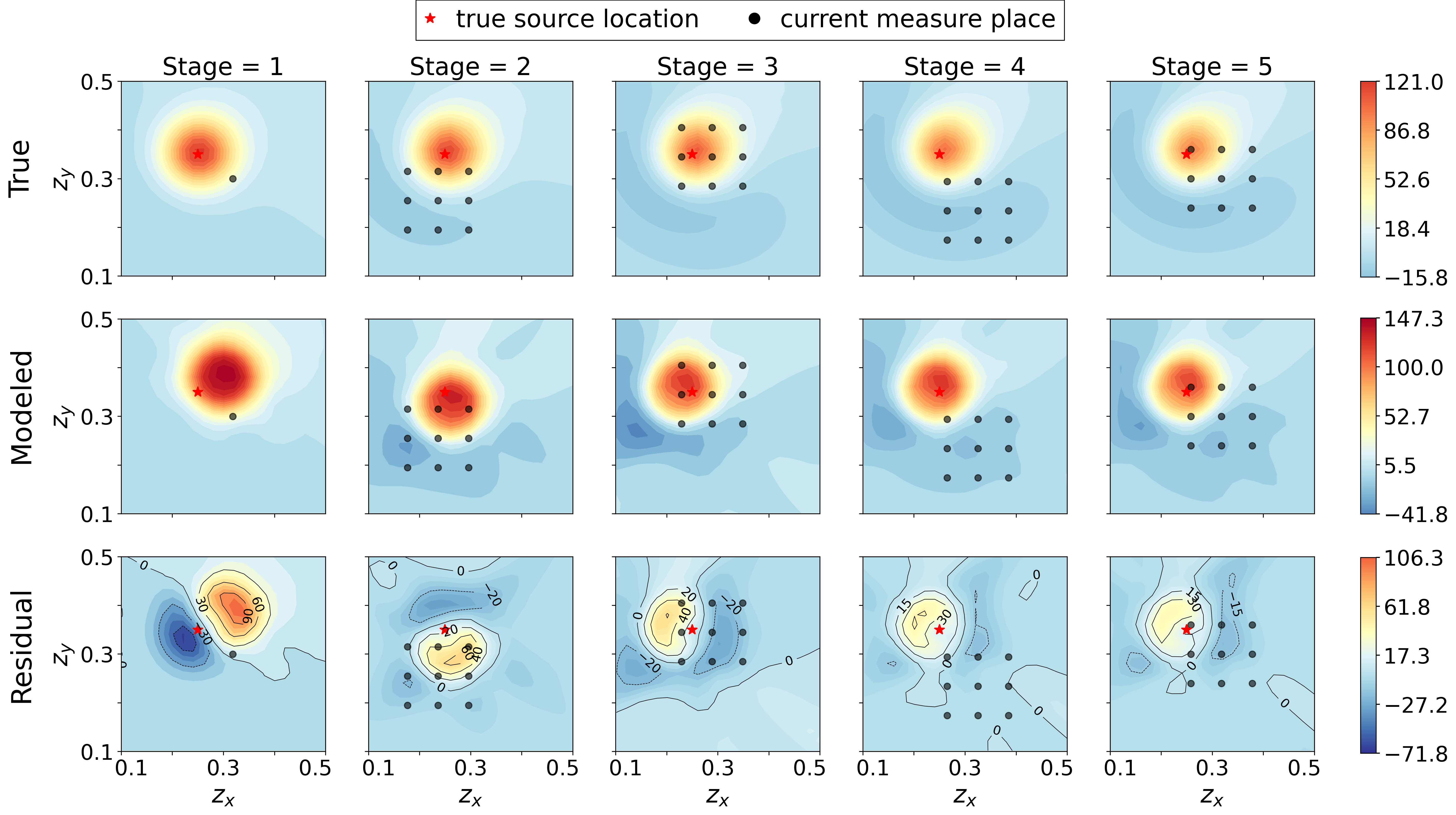}
    \caption{Results for the RHS of the PDE}
    \label{fig: rhs re}
  \end{subfigure}%
  \vspace{10pt}
  \begin{subfigure}[b]{1\textwidth}
    \centering
    \includegraphics[width=0.95\linewidth]{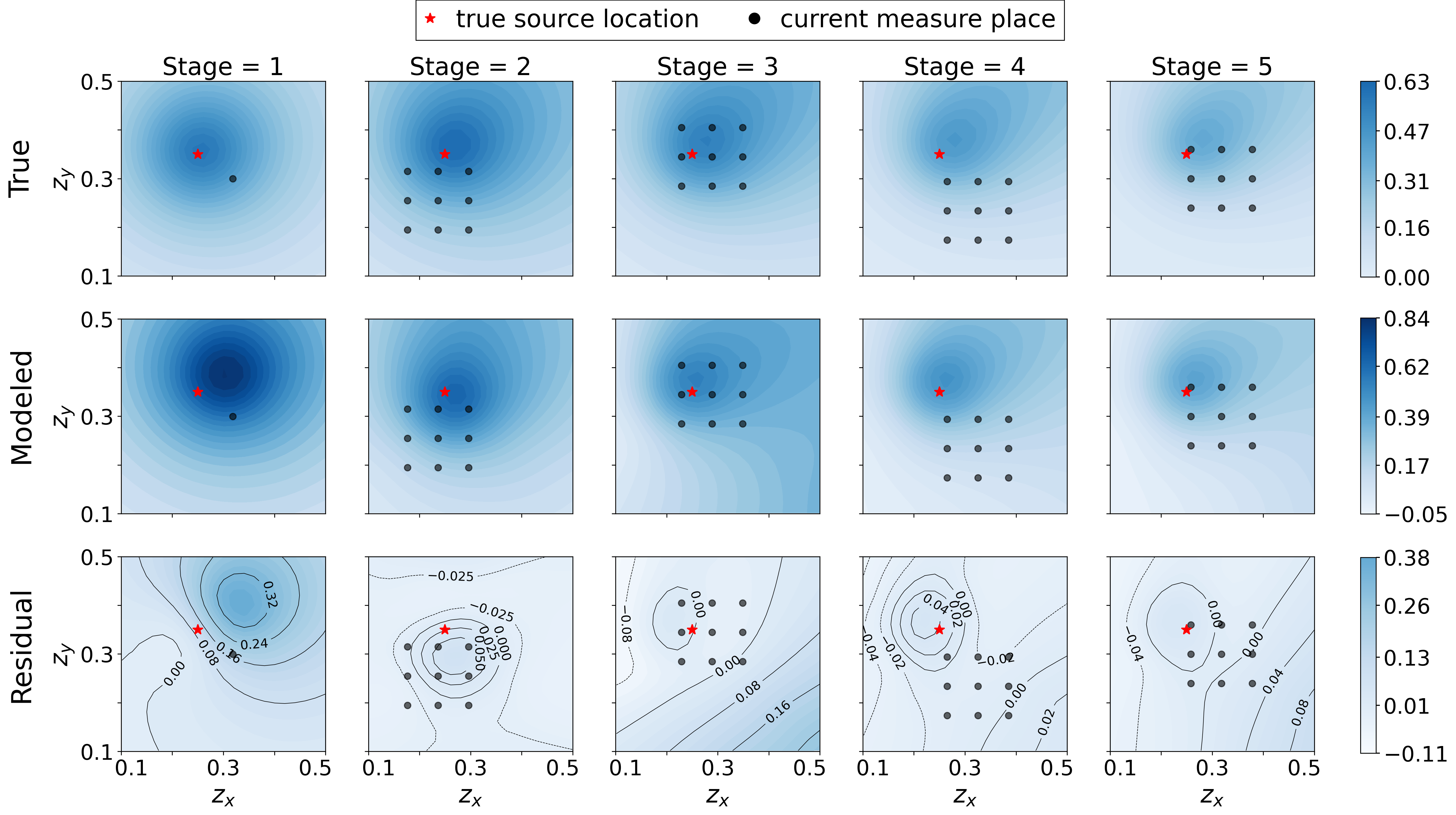}
    \caption{Results for the solution field of the PDE}
    \label{fig: u re}
  \end{subfigure}%
  \caption{Comparison between the true data, modeled approximation, and residuals across five measurement stages for RHS and solution field of the PDE: (a) Results for the RHS of the PDE, and (b) Results for the solution field of the PDE. Each panel consists of three rows: the first row shows the true data, the second row displays the modeled approximation, and the third row presents the residuals (difference between the true and modeled values). The contour plots are zoomed into the region $[0.1,0.5]^2$, which corresponds to the primary data collection area for model calibration. Within each panel, identical colors represent the same numerical values throughout the entire figure. For modeled results, stage 1 shows the initial model before any update, while subsequent stages present the updated model at each corresponding stage. In the first stage, model correction is skipped, and only the design point for the physical parameter is shown.}
  \label{fig: correct network error 3}
\end{figure}

We further investigate the detailed results of the PDE RHS (excluding the diffusion term) and the solution field across successive update stages. In Fig.~\ref{fig: rhs re}, the results indicate that the modeled right-hand side is gradually matching the true system within the collected-data region. Consequently, in Fig.~\ref{fig: u re}, the solutions of the modeled system (second row) gradually converge to the true field (first row), and the overall residual is steadily reduced, demonstrating that the solution field is effectively corrected. This confirms that the neural-network-corrected RHS remains close to the ground truth, yielding accurate posterior inferences overall, although $k_y$ does not correctly converge to the value taken by the true system.

These results confirm that our approach effectively aligns the modeled system with the true one. This neural-network-corrected modeled system thus ensures an accurate and robust posterior estimation of the physics-based parameters treated as random variables and updated via Bayesian methods. It should be noted that treating some of the physics-based parameters deterministically and optimizing them jointly with neural network parameters can cause the neural network to compensate the incorrect values taken by those physics-based parameters, impairing the identifiability of those parameters and leading to a challenging ill-posed problem.

\section{Discussions}
\label{sec:discussions}

This work adopts a Bayesian framework rooted in information theory while aligning more closely with traditional experimental design due to the reliance on measured data. Specifically, we conduct preliminary experiments with several proposed designs, evaluate the information gained from each, and use this information to optimize for an improved design. While the likelihood and posterior updates rely on an inaccurate model, the results presented in Section~\ref{sec: Numerical Results} demonstrate that this approach yields highly informative designs with respect to effectively enabling model correction despite discrepancies. The optimization problem can be practically addressed by proposing several candidate designs and selecting or interpolating the most informative one. Alternatively, it can be solved using either gradient-based methods or derivative-free optimization techniques. It is worth noting that the numerical results of Section~\ref{sec: Numerical Results} only employ the measurement data of the optimal design point for model error correction, instead of the measurement data on a trajectory of points that lead to the optimal design point. Admittedly, more data along the whole optimization trajectory tends to provide more information gain, while the calibration of the model error can possibly be worse by using the trajectory of data points, and thus the criterion based on information gain to assess the data informativeness on correcting the model error needs to build on a fixed amount of data points, e.g., a single data point in this work. The reason for only adopting the data at the optimal design point for a given stage is mainly due to the direct contribution of the optimal design point in the posterior distribution estimation of the subsequent stage. Therefore, a better match of measurement data on the optimal design point is more favorable than an overall good match of measurement data on the whole optimization trajectory, for which the model accuracy on the optimal design point could be compensated a bit to ensure the overall agreement with the whole trajectory data.

A key consideration in hybrid models that combine physics-based components with neural-network-based correction terms is the ill-posedness, where the network may compensate or even replace existing physics-based components, thereby obscuring the actual meaning of those physics-based parameters being estimated. In certain cases, such as the numerical example studied in Section~\ref{sec:Correct functional error}, such ill-posedness can be mitigated by imposing desired constraints on the trained neural network, e.g., by enforcing the prescribed center-decay pattern with assigned center and restricting the network to spatial variation only, while the time-varying convective term is handled entirely by the physics-based model. More broadly, general cases may require other strategies like promoting orthogonality or post-training decomposition. An example of post-training decomposition is a re-attribution step to find an adjusted estimate $\boldsymbol{\theta}_\mathcal{G}^\text{adj}$ by requiring the physical model $\mathcal{G}(\mathbf{u};\boldsymbol{\theta}_\mathcal{G})$ to maximally explain the learned RHS $H(\mathbf{u})=\mathcal{G}(\mathbf{u};\boldsymbol{\theta}_\mathcal{G}^*)+\text{NN}(\mathbf{u};\boldsymbol{\theta}_\text{NN})$, via solving an additional optimization problem $\boldsymbol{\theta}_\mathcal{G}^\text{adj}=\arg \max_{\boldsymbol{\theta}_\mathcal{G}} \Vert \mathcal{G}(\mathbf{u};\boldsymbol{\theta}_\mathcal{G})-H(\mathbf{u}) \Vert^2 _2$. The $\boldsymbol{\theta}_\mathcal{G}^\text{adj}$ is thus an updated estimation of the physics-based parameters where all aspects of the trained RHS explainable by the physics-based model $\mathcal{G}$ are attributed to it. Although this post-training decomposition is not investigated in this work, it leads to a meaningful future direction for the joint learning of physics-based parameters and structural errors.

In the numerical results of Section~\ref{sec: Numerical Results}, we mainly focused on the scenario that allows some experimental measurements during the iterative stages of sequential experimental design. We have also tested the scenario that solely depends on the numerical model simulations, and the comparisons indicate that both predicted and measured data are effective within our framework. The use of simulation data yields less satisfactory performance and demands further technical refinement. The reason for this performance gap is that the EIG of a given design can differ significantly from the actual information gain it provides. When the model is correct, this difference is primarily determined by the accuracy of the prior information. However, in the presence of model discrepancy, errors in the likelihood function further amplify this difference, leading to a scenario where the design identified based on EIG may yield incorrect information gain. The biased likelihood function is utilized twice in the computation of basic version EIG: (i) to calculate the predicted distribution of measurements, and (ii) to update the posterior distribution after obtaining a specific prediction. By using real measurements, the first step can be bypassed, thereby mitigating the impact of model inaccuracies. Consequently, the computed information gain represents a more reliable metric of information provided by the data. To demonstrate the strength of the hybrid correction framework, we focus on using measurement data in our design process and leave the scenario that solely depends on numerical model simulations to approximate EIG for future work.

In our framework, we partition the parameters into two groups and handle each group with a distinct update mechanism, which raises the question of how to select a representative point estimate from a probabilistic posterior for use in the deterministic gradient updates. We address this by performing multiple gradient-based optimizations to find distinct local maxima, a strategy that allows us to more effectively synthesize information from the posterior landscape. We then use the mean of these identified peaks as the averaged point estimate for the physics-based parameters. We have demonstrated that this strategy is both effective and robust when the true posterior is a unimodal distribution, or at least dominated by a leading mode. If the true posterior of the physics-based parameters is apparently multimodal, averaging of samples may yield an estimate that lies between peaks and degrades network performance at each mode. In such cases, it is potentially advantageous to train different neural networks at each representative mode and then weight their contributions by distance or likelihood, which is a meaningful extension for future study.

\section{Conclusion}\label{Conclusion}
In this work, we propose an efficient hybrid approach to actively learning the model discrepancy of a digital twin based on the data from the sequential BED. We first validate the performance of the proposed method in a classical numerical example governed by a convection-diffusion equation, for which full BED is still feasible. The results confirm that the proposed method can effectively calibrate the low-dimensional parametric model error based on the optimal designs from the BED. The proposed method is then further studied in the same numerical example with a high-dimensional structural model error, for which full BED is not practical anymore. It is worth noting that this numerical example could suffer from the ill-posedness issue, which stems from jointly learning the physics-based model parameters and the neural-network-based correction term. Such an ill-posedness issue is mitigated in this work by restricting the network to learn only the spatial discrepancy, while delegating all time-dependent terms to the physics-based model. We observe that most of the optimal designs from the BED that focus on the low-dimensional physics-based parameters can still be informative for calibrating the model discrepancy, while some of those designs could gradually lead to a larger model discrepancy. Therefore, we propose an ensemble-based approximation of information gain to assess the data informativeness and to enhance the active learning model discrepancy. The results show that the proposed method is efficient and robust to active learning of the high-dimensional model discrepancy. With the promising results of the hybrid approach demonstrated in this work, some future directions are summarized below:
\begin{itemize}
    \item Using the ensemble-based approximation of information gain to identify the optimal designs and informative data that are directly applicable to the calibration of model discrepancy.
    \item Extending the ensemble-based approximation of information gain to handle non-Gaussian prior and posterior distributions.
\end{itemize}

\section*{Acknowledgments}
H.Y., C.C., and J.W. are supported by the University of Wisconsin-Madison, Office of the Vice Chancellor for Research and Graduate Education with funding from the Wisconsin Alumni Research Foundation.

\section*{Data Availability}
The codes and data that support the findings of this study are available in the link: \url{https://github.com/AIMS-Madison/BED-Model-Discrepancy}. 
  
\bibliographystyle{unsrt}
\bibliography{references}

\clearpage
\appendix

\section{Theoretical global maximum}
\label{appendx proof for obj in model correction}

The model discrepancy $\mathcal{G}^\dagger - \mathcal{G}$  which is characterized by a neural network $\textrm{NN}(\mathbf{u}; \boldsymbol{\theta}_{\textrm{NN}})$ in Eq.~\eqref{eq:modeled_system} can be calibrated by solving the optimization problem in Eq.~\eqref{eq:optimal_theta_NN} where the objective function $L(\boldsymbol\theta_\text{NN};\ \boldsymbol\theta_\mathcal{G}^*,\mathbf{y},\mathbf{d})$ is defined as $p(\mathbf{y}|\boldsymbol\theta_\mathcal{G}^*,\mathbf{d};\boldsymbol\theta_\text{NN})$ which is the likelihood function of $\boldsymbol{\theta}_{\textrm{NN}}$. The solution obtained by maximizing the likelihood function is the optimal solution for the modeled system to approximate the true system. The universal approximation property of neural networks guarantees the existence of a solution to this equation. We further demonstrate why this objective function leads to the solution.

The likelihood function is written as follows: 
\begin{equation}
\begin{aligned}
    p(\mathbf{y} |\boldsymbol\theta^*_{\mathcal{G}},\mathbf{d};\boldsymbol\theta_\text{NN}) = \frac{1}{\sqrt{|2\pi \boldsymbol{\Sigma}_{\boldsymbol{\epsilon}}|}}\exp\left(-\frac{1}{2}  \boldsymbol{\Sigma}_{\boldsymbol{\epsilon}}^{-1}\left\Vert\mathbf{y} - \mathbf{u}(\mathbf{d};\boldsymbol\theta^*_{\mathcal{G}},\boldsymbol\theta_\text{NN}) \right\Vert^2_2 \right) 
\end{aligned}
\end{equation}
where $\mathbf{y}=\mathbf{u}^\dagger(\mathbf{d};\boldsymbol\theta^\dagger)+\boldsymbol{\epsilon}$ and $\mathbf{u}^\dagger$ indicates the true field from system.

Maximizing this likelihood function is equivalent to minimizing the negative log-likelihood, which can further be reformulated as minimizing the weighted mean squared error between $\mathbf{y}$ and $\mathbf{u}(\mathbf{d}; \boldsymbol{\theta}_\mathcal{G}^*, \boldsymbol{\theta}_\text{NN})$. Given sufficient data:
\begin{itemize}
    \item The measurement error $\boldsymbol{\epsilon}$ can be averaged out to enable the predictive field $\mathbf{y}$ to approximate the true field $\mathbf{u}^\dagger(\mathbf{d}; \boldsymbol{\theta}^\dagger)$;
    \item The pointwise alignment between $\mathbf{y}$ and $\mathbf{u}(\mathbf{d}; \boldsymbol{\theta}_\mathcal{G}^*, \boldsymbol{\theta}_\text{NN})$ ensures that the true field $\mathbf{u}^\dagger$ aligns with the simulated field $\mathbf{u}$.
\end{itemize}

This alignment at the field level further ensures the dynamic consistency between the true system $\mathcal{G}^\dagger$ and the modeled system $\mathcal{G} + \textrm{NN}$:
\[
\mathcal{G}^\dagger(\mathbf{u}; \boldsymbol{\theta}^\dagger) = \mathcal{G}(\mathbf{u}; \boldsymbol{\theta}_\mathcal{G}^*) + \textrm{NN}(\mathbf{u}; \boldsymbol{\theta}_\text{NN}).
\]

With $\boldsymbol{\theta}_\mathcal{G}^*$ approximating $\boldsymbol{\theta}_\mathcal{G}^\dagger$, a result progressively provided by Bayesian OED\cite{lindley_bayesian_1972,sebastiani_maximum_2000}, the above alignment equation can represent the equivalence between Eq.~\eqref{eq:true_system} and Eq.~\eqref{eq:modeled_system}. The universal approximation property of neural networks guarantees the existence of a solution, which corresponds to the optimized $\boldsymbol{\theta}_{\text{NN}}^*$.

However, in the settings of traditional BED, $\boldsymbol{\theta}_{\text{NN}}^*$ can not fully capture the model discrepancy $\mathcal{G}^\dagger - \mathcal{G}$ when $\boldsymbol{\theta}_\mathcal{G}^*$ does not approximate $\boldsymbol{\theta}_\mathcal{G}^\dagger$. Through the method proposed in this paper including the iterative process of BED and model correction, both $\boldsymbol{\theta}_\mathcal{G}^*$ and $\boldsymbol{\theta}_{\text{NN}}^*$ can progressively converge to their respective true values.

Therefore, the choice of likelihood function as the objective to train the modeled system can capture the model discrepancy under the following conditions:  
\begin{itemize}
    \item Multiple experiments can average measure error $\boldsymbol{\epsilon}$ out;
    \item $\boldsymbol\theta_\mathcal{G}^\dagger$ can be approximated by $\boldsymbol\theta^*_\mathcal{G}$ from the its posterior by BED;
    \item The alignment of the fields ensures that the modeled system $\mathcal{G}+\text{NN}$ captures the dynamics of the true system $\mathcal{G}^\dagger$.
\end{itemize}

\section{Detailed derivation of the Jacobian matrix}
\label{sec:Jacobian_matrix}

We follow the PDE in Eq. \eqref{eq: c sourse} but assume the PDE solver is not auto-differentiable. To facilitate the adjoint method for the calculation of $\mathrm{d} \text{Obj}/\mathrm{d} \boldsymbol{\theta}_\text{NN}$, we first compute the Jacobian matrix $\partial (\mathcal{G}+\textrm{NN})/\partial \mathbf{u}$ using the functional derivative method \cite{courant_methods_1954}\cite{han_theoretical_2009} from the analytical PDE. In addition, we introduce other potential methods for obtaining the Jacobian matrix at the end of this section and present the results of the adjoint method using these different Jacobians in the next section.

The scalar field of model state $\mathbf{u} \in \mathbb{R}^{N_\mathbf{u} \times N_\mathbf{u}}$ in the PDE Eq. \eqref{eq: c sourse} is represented as a flattened vector $\mathbf{u} \in \mathbb{R}^{N_\mathbf{u}^2}$ and its corresponding Jacobian matrix is of shape $N_\mathbf{u}^2 \times N_\mathbf{u}^2$. We start by considering the $i$th element of the flattened $\mathbf{u}$:
\begin{equation}
    \frac{\partial \mathbf{u}_i(\mathbf{z};\boldsymbol\theta)}{\partial t}=\mathcal{G}_i(\mathbf{u};\boldsymbol\theta)+\textrm{NN}_i(\mathbf{z};\boldsymbol\theta).
\end{equation}

The functional derivative of $\mathcal{G}_i+\textrm{NN}_i$ with respect to $\mathbf{u}$ is

\begin{equation}
    \begin{aligned}
    \sum_{\mathbf{z} \in \mathcal{Z}} \frac{\delta (\mathcal{G}_{i}+\textrm{NN}_i)}{\delta \mathbf{u}}(\mathbf{z})\eta(\mathbf{z}) &= \left. \frac{\mathrm{d}\mathcal{G}(\mathbf{u} + h\eta)_{i}}{\mathrm{d}h} \right|_{h=0}+\underset{\mathclap{=0~\text{(the network is not explicit function of $\mathbf{u})$}}}{\underbrace{\left. \frac{\mathrm{d}\textrm{NN}(\mathbf{u} + h\eta)_{i}}{\mathrm{d}h} \right|_{h=0}}}\\
    &= \left. \frac{\mathrm{d}}{\mathrm{d}h} \left\{ \nabla^2(\mathbf{u} + h\eta(\mathbf{z})) -v(t)\cdot\nabla (\mathbf{u} + h\eta(\mathbf{z})) \right\}_{i} \right|_{h=0} \\&~~~~\text{(the forcing terms are not explicit function of $\mathbf{u})$}\\
    &= \left.\left\{\nabla^2 \eta(\mathbf{z}) -v(t)\cdot\nabla \eta(\mathbf{z})\right\}_{i}\right|_{h=0} \\
    &= \left\{\left[\nabla^2  -v(t)\cdot\nabla \right]\eta(\mathbf{z})\right\}_{i} \\
    &=\sum_{\mathbf{z} \in \mathbf{Z}}e_i^T\cdot\left[\nabla^2  -v(t)\cdot\nabla \right]\eta(\mathbf{z}).
    \end{aligned}\label{eq:functional derivative defination}
\end{equation}
where $e_i$ represents the standard basis vector in the $i$-th direction, used to extract the $i$-th component of the vector field. 

It should be noted that in Eq. \eqref{eq: c sourse}, network term $\textrm{NN}$ is not a function of model state $\mathbf{u}$ so we directly cancel the term $ \mathrm{d}\mathcal{G}(\mathbf{u} + h\eta)_{i}/\mathrm{d}h$ in Eq. \eqref{eq:functional derivative defination}. If in other cases the network term is a function of model state $\mathbf{u}$, this term can be easily obtained by the auto-differentiability of neural networks.

The functional derivative result  $\frac{\delta \mathcal{G}_i+\textrm{NN}_i}{\delta \mathbf{u}}(z)=e_i^T\cdot\left[\nabla^2  -v(t)\cdot\nabla \right]$ forms the $i$-th row of the Jacobian matrix.  The specific value for discretized numerical tests depends on the employed finite difference form. 

Considering the 2D central difference of $\nabla^2_h$ ($h$ denotes the discretized form) on the flatten scalar field $\mathbf{u}\in \mathbb{R}^{N_\mathbf{u}^2}$, the $i$-th element can be expressed as:
\begin{equation}
    (\nabla^2_h \mathbf{u})_{i} \approx \frac{1}{h^2}(u_{(i+N_\mathbf{u})}+u_{(i-N_\mathbf{u})}+u_{(i+1)}+u_{(i-1)}-4u_{i})=\frac{1}{h^2}\mathbf{A}_i \cdot \mathbf{u},
\end{equation}
where $\mathbf{A}_i \in \mathbb{R}^{N_\mathbf{u}^2}$ is

\begin{equation}
    (\mathbf{A}_i)_j=\left\{\begin{array}{rl}
         1,&  j=i\pm1,i\pm N_\mathbf{u})\\
         -4,& j=i\\
         0,& \text{otherwise}
    \end{array}\right.
\end{equation}

Following this formula, the whole Jacobian matrix $\mathbf{A}$ of Laplace operator $\nabla^2_h$ can be expressed as:
\begin{equation}
    \mathbf{A} =\frac{1}{h^2} \left[\begin{matrix}
        \mathbf{T} & \mathbf{I}& \mathbf{0} & \cdots & \mathbf{0}\\
        \mathbf{I} & \mathbf{T}& \mathbf{I} & \cdots & \mathbf{0}\\
        \mathbf{0} & \mathbf{I}& \mathbf{T} & \cdots & \mathbf{0}\\
        \vdots & \vdots& \vdots & \ddots & \mathbf{I}\\
        \mathbf{0}& \mathbf{0}& \mathbf{0} & \mathbf{I} & \mathbf{T}
    \end{matrix}\right]_{N_\mathbf{u}^2\times N_\mathbf{u}^2},
\end{equation}
where $\mathbf{A} \in \mathbb{R}^{N_\mathbf{u}^2\times N_\mathbf{u}^2}$ is composed of $N_\mathbf{u} \times N_\mathbf{u}$ blocks, specifically $\mathbf{I}$, $\mathbf{0}$, and $\mathbf{T}$, all of which are matrices in $\mathbb{R}^{N_\mathbf{u} \times N_\mathbf{u}}$. In detail, $\mathbf{I}$ is the $N_\mathbf{u}\times N_\mathbf{u}$ identity matrix, $\mathbf{0}$ is the $N_\mathbf{u}\times N_\mathbf{u}$ zero matrix, and $\mathbf{T}$ is
\begin{equation}
    \mathbf{T}=\left[\begin{matrix}
        -4&1&0&\cdots&0\\
        1&-4&1&\cdots&0\\
        0&1&-4&\cdots&0\\
        \vdots&\vdots&\vdots&\ddots&1\\
        0&0&0&1&-4\\
    \end{matrix}
    \right]_{N_\mathbf{u}\times N_\mathbf{u}}.
\end{equation}

The above difference scheme does not include specific modifications at the boundaries. Any necessary adjustments to the boundary points depend on the specific configuration of the computational case and the boundary conditions being applied.

For the same reason, the Jacobian matrix $\mathbf{B}$ of advection term $v(t)\cdot \nabla$ in 2D central difference form is
\begin{equation}
    \mathbf{B} =\frac{1}{2h} \left[\begin{matrix}
        \mathbf{X} & \mathbf{Y}& \mathbf{0} & \cdots & \mathbf{0}\\
        -\mathbf{Y} & \mathbf{X}& \mathbf{Y} & \cdots & \mathbf{0}\\
        \mathbf{0} & -\mathbf{Y}& \mathbf{X} & \cdots & \mathbf{0}\\
        \vdots & \vdots& \vdots & \ddots & \mathbf{Y}\\
        \mathbf{0}& \mathbf{0}& \mathbf{0} & -\mathbf{Y} & \mathbf{X}
    \end{matrix}\right]_{N_\mathbf{u}^2\times N_\mathbf{u}^2},
\end{equation}
where $\mathbf{B} \in \mathbb{R}^{N_\mathbf{u}^2\times N_\mathbf{u}^2}$ is composed of $N_\mathbf{u} \times N_\mathbf{u}$ blocks, specifically $\mathbf{X}$, $\mathbf{Y}$, and $\mathbf{0}$, all of which are matrices in $\mathbb{R}^{N_\mathbf{u} \times N_\mathbf{u}}$. In detail:
\begin{equation}
     \mathbf{X}=v_x\left[\begin{matrix}
        0&1&0&\cdots&0\\
        -1&0&1&\cdots&0\\
        0&-1&0&\cdots&0\\
        \vdots&\vdots&\vdots&\ddots&1\\
        0&0&0&-1&0\\
    \end{matrix}
    \right]_{N_\mathbf{u}\times N_\mathbf{u}},  \quad  \mathbf{Y}=v_y\mathbf{I},\quad v(t)=(v_x,v_y) .
\end{equation}

The current form is derived under the numerical case where $v(t)$ is spatially independent as discussed in this paper. More complex form of the Jacobian matrix if $v(t)$ depends on spatial position could refer to any numerical method textbook.

The whole desired Jacobian matrix for Eq.~\eqref{eq:true_system_example} equals to $\mathbf{A}-\mathbf{B}$. Higher-order difference forms primarily alter the values of specific elements and those in their immediate vicinity. However, the sum of these vectors should invariably equal one. The specific finite difference scheme to pick should align with the forms adopted by the PDE solver.

\section{Additional results of an acoustic inverse problem}
\label{apd: Numerical case: acoustic inverse}

The wave equation:
\begin{equation}
    \frac{\partial^2 u(\mathbf{r},t)}{\partial t^2}-c^2\nabla^2u(\mathbf{r},t)=0, \quad \mathbf{r} \in \Omega
\end{equation}
describes how a physical quantity propagates through space and time. For the steady-state or single-frequency case, we assume that the solution has a harmonic time dependence. By substituting an ansatz of the form
\begin{equation}
    u(\mathbf{r},t)=U(\mathbf{r})e^{i\omega t},
\end{equation}
into the wave equation, the time-dependent part factors out. This process leads to Helmholtz equation for the spatial component:
\begin{equation}
    \nabla^2 U(\mathbf{r})+k^2U(\mathbf{r})=0,
\end{equation}
where the wave number $k=\omega / c=2\pi/\lambda$, with $c$ represents the phase speed and $\lambda$ represents the wave length. The complex acoustic pressure amplitude function $U(\mathbf{r})$ is complex-valued, which represents both the amplitude and phase information at each point $\mathbf{r}$. The Helmholtz equation is central to understanding the spatial behavior of waves under monochromatic conditions, making it invaluable in fields such as acoustics, electromagnetics, and quantum mechanics.

\begin{figure}[H]
  \centering
  \begin{subfigure}[b]{0.5\textwidth}
    \centering
    \includegraphics[width=0.95\linewidth]{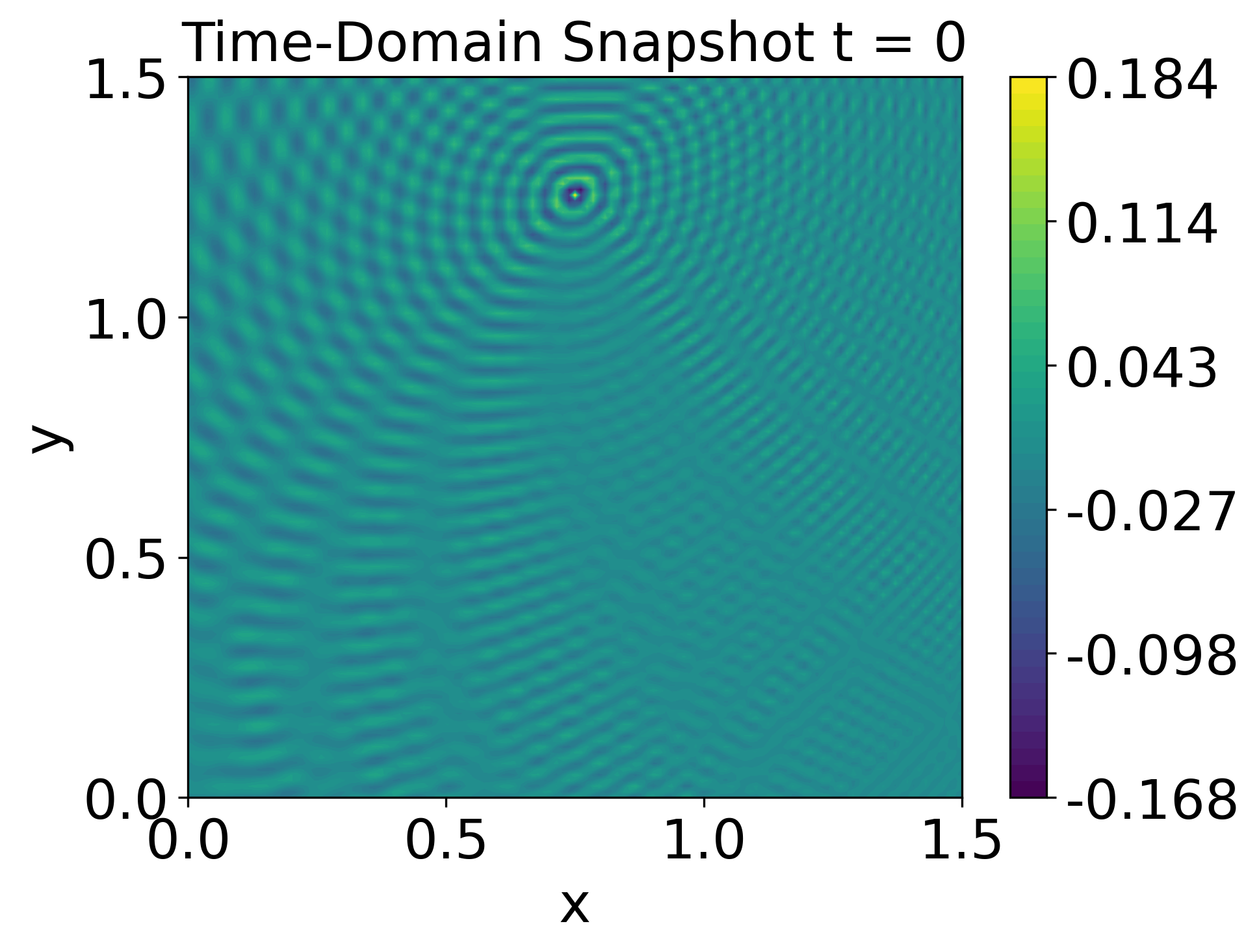}
    \caption{Example wave field at time unit 0}
  \label{fig:Example acoustic field at time unit 0}
  \end{subfigure}%
  \begin{subfigure}[b]{0.5\textwidth}
    \centering
    \includegraphics[width=0.95\linewidth]{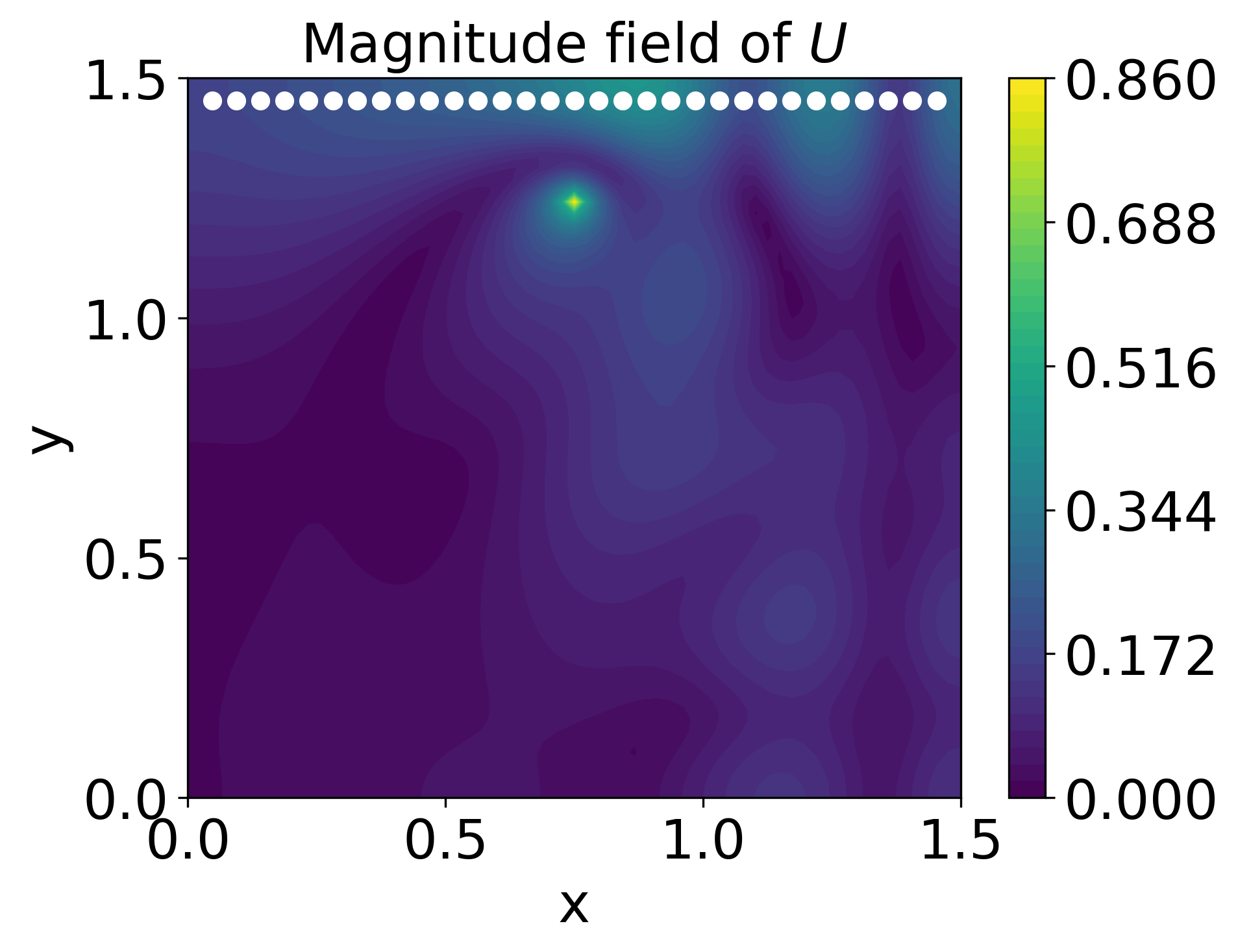}
    \caption{True magnitude field of the complex amplitude}
  \label{fig:True magnitude field of the complex amplitude}
  \end{subfigure}
  \caption{Field results of the acoustic system. Panel (a) is the time-domain snapshot of wave field at time unit 0 with k=100. Panel (b) is the magnitude of the complex amplitude $U$ with k=5, which is true field in the inverse problem. The white dots denote the potential measure places.}
  \label{fig: field results of acoustic case}
\end{figure}

We consider the problem of a point source acoustic wave propagating in a two-dimensional inhomogeneous medium~\cite{wu_large-scale_2023, oleary-roseberry_learning_2022}:
\begin{equation}
    \begin{aligned}
        -\nabla^2 U(x,y)-e^{2m(x,y)}k^2U(x,y)&=f, \quad \{x,y\}\in \Omega,\\
        f&=A\delta (x-x_0,y-y_0),\\
        \nabla u \cdot \mathbf{n}&=0 ~\text{on}~ \Gamma_\text{top},\\
        \text{PML boundary condition}~&~\text{on }~\partial \Omega/\Gamma_\text{top}.
    \end{aligned}
\end{equation}
where $e^{2m(x,y)}$ represents the spatially varying medium property, such as the local reflection coefficient or refractive index contrast. $f$ represents a time-harmonic point source of amplitude $A$, located at $\{x_0,y_0\}$ and modeled by Dirac delta function $\delta$. The computational domain $\Omega$ is a square region, subject to a reflective (Neumann) boundary condition on the top boundary $ \Gamma_\text{top}$ and surrounded by perfectly matched layers (PML) on the remaining three boundaries to simulate wave absorption. Due to the inhomogeneity, the medium’s physical properties vary with position, which leads to complex propagation behavior. Increasing $m(x,y)$ reduces the local wave speed and increases the effective refractive index, resulting in stronger reflection and denser wavefronts with more rapid spatial oscillations. For example, Fig.~\ref{fig:Example acoustic field at time unit 0} shows the time-domain wave field snapshot of the case with $m(x,y)=2/3 x$, a high wave number $k=100$, and a source located in $\{x_0,y_0\}=\{0.75,1.25\}$, where the wavefronts to become progressively denser in regions with larger $x$-coordinate. Also, the reflection can be clearly seen near the top boundary.

\textbf{Problem setup}: Assume the unknown physical parameter is the source amplitude $A$ with true value $1.5$, and a model discrepancy lies in the medium properties. The true form of $m^\dagger (x,y)$ in system is given by
\begin{equation}
    m^\dagger(x,y)=\frac{2}{3} x,
\end{equation}
and the modeled $m(x,y)$ is
\begin{equation}
    m(x,y)=\frac{2}{3}x^2.
\end{equation}
A neural network (denoted by $\text{NN}$) is utilized to capture the discrepancy between the modeled and true medium properties: $\text{NN} \rightarrow \frac{2}{3}(x-x^2)$. The network takes the spatial coordinate $x$ as input and outputs a scalar medium parameter $m$ at spatial location $\{x,y\}$: $\text{NN}: x \mapsto m$. The architecture is a fully connected feed-forward network featuring a single hidden layer with five nodes, which results in a total of 16 parameters $\boldsymbol{\theta}_\text{NN} \subseteq \mathbb{R}^{16}.$

 The general goal of this case is to seek a design $\mathbf{d}$ as where to measure $U$ and use the measurement $\mathbf{y}=U(\mathbf{d})+\epsilon,~\epsilon\sim \mathcal{N}(0,\sigma^2)$ to infer the source amplitude $A$. A brief procedure for each sequential experiment is described by
 \begin{itemize}
     \item Find optimal $\mathbf{d}$ for $A$ by BED method.
     \item Observe $\mathbf{y}$ and update a posterior for $A$: $p(A|\mathbf{y},\mathbf{d})$ with a MAP estimation $A^*$.
     \item Optimize the network parameter $\boldsymbol{\theta}_\text{NN}$ by maximizing likelihood $p(\mathbf{y}|\mathbf{d},A^*,\boldsymbol{\theta}_\text{NN})$.
     \item Use the updated model to re-update a posterior $p(A|\mathbf{y},\mathbf{d})$.
 \end{itemize}
The posterior of each experiment will be utilized as the prior for the next experiment. And the updated model is also passed into the next experiment.

\textbf{Numerical Setting and Solver}: The computational domian $\Omega$ is defined as $[0,1.5]^2$, discretized uniformly into $64\times 64$ cells. The wave number is set to $k=5.0$ and the source is located at $\{x_0,y_0\}=\{0.75,1.25\}$. The potential measure places are constrained to the set $\{\frac{3}{128}+\frac{3}{64}i,\frac{189}{128}\}_{i=0}^{31}$ as shown in Fig.~\ref{fig:True magnitude field of the complex amplitude}. This setup follows the cases in~\cite{wu_large-scale_2023, oleary-roseberry_learning_2022}, but adopts a scaled-down version. The scaling is intentional: while the referenced works focus on tractable estimation of high-dimensional expected information gain, our emphasis lies in learning and correcting model discrepancy. Despite the simplified scaling, the simulated wave fields closely resemble those presented in the original works under the same setting. The prior distribution of $A$ is assumed to be uniform over $[0,3]$, discretized into 151 equally spaced points. A finite element method solver is developed entirely in JAX for its automatic differentiation capabilities. The demonstration example will be available in our GitHub repository.

\begin{figure}[H]
    \centering
    \includegraphics[width=0.85\linewidth]{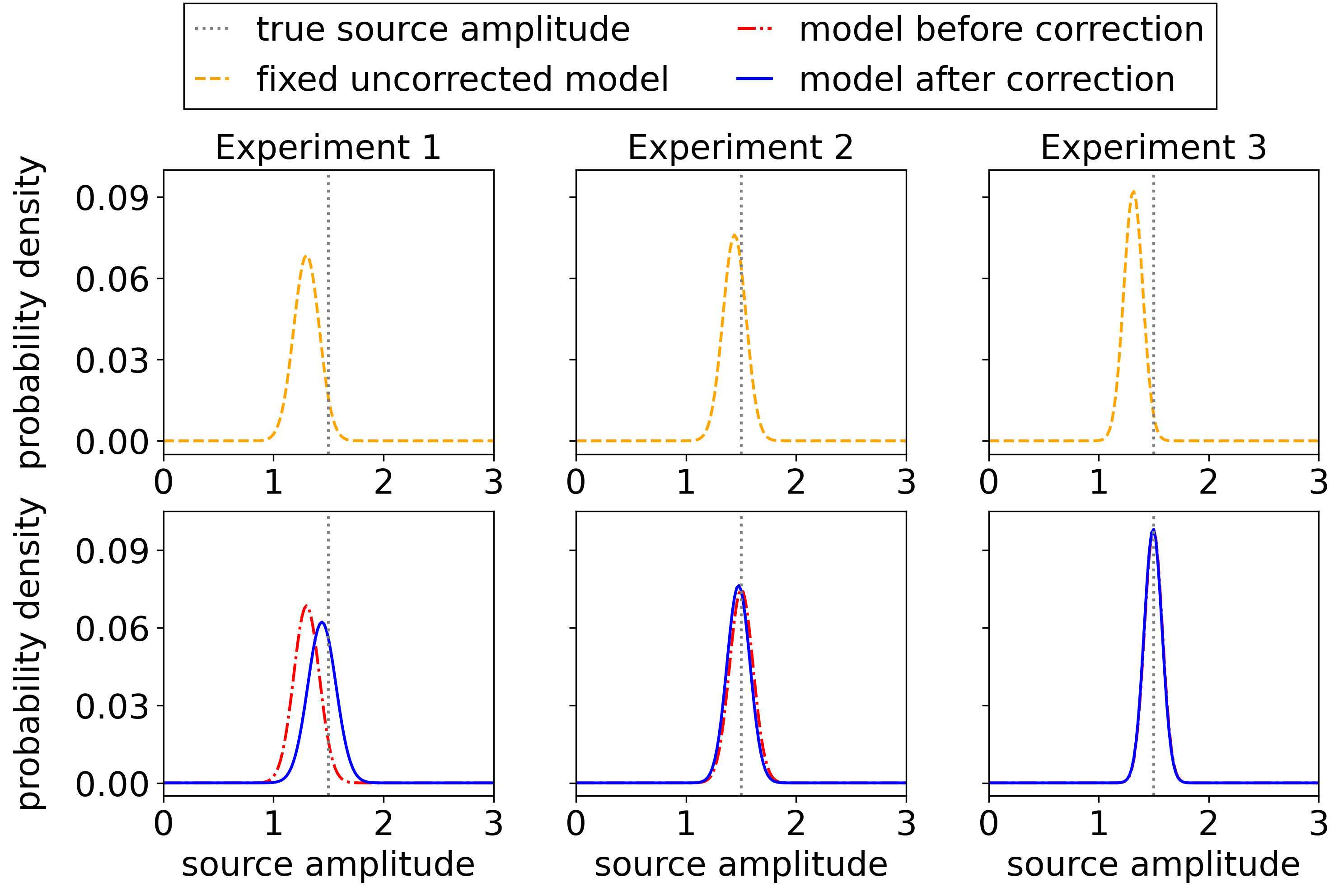}
    \caption{Posterior distribution of source amplitude $A$}
    \label{fig:Posterior distribution of source amplitude}
\end{figure}

\begin{figure}[H]
    \centering
    \includegraphics[width=0.45\linewidth]{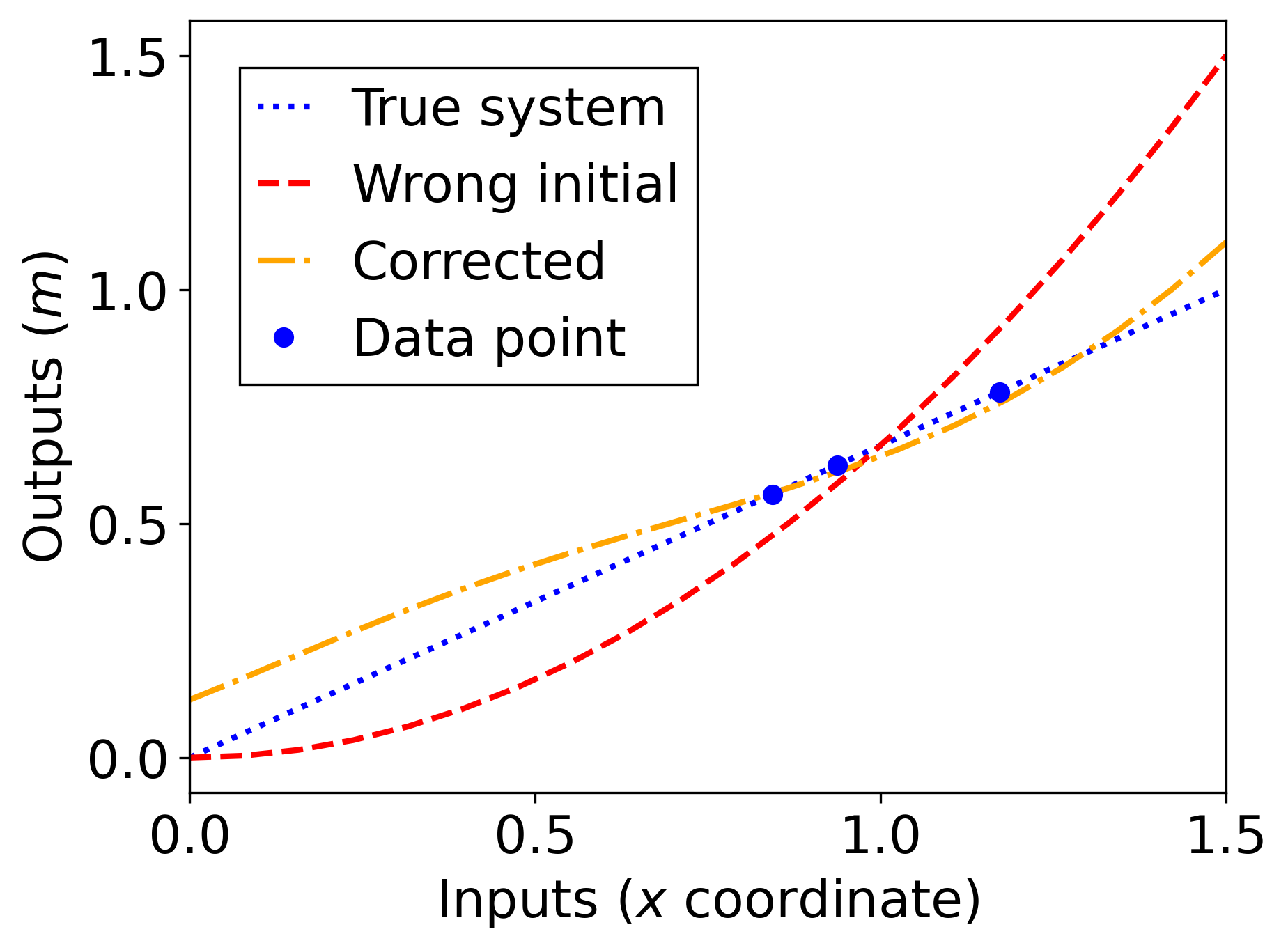}
    \caption{Comparison of Medium Property Fields: True $m^\dagger(x)$, Initial $m(x)$ and Corrected $m(x)+\text{NN}(x|\boldsymbol{\theta}_\text{NN})$. The data points represents the $x$-coordinates corresponding to the optimal design along with the associated true values $m^\dagger$. However, it is important to note that the data used in model correction are not these $\{x,m^\dagger\}$ pairs; rather, they are the complex amplitude measurements obtained at the design coordinates.}
    \label{fig:Network-corrected model performance}
\end{figure}

\textbf{Results}: We begin by applying the standard BED to find design and inference source amplitude using an uncorrected model. The results, presented in the first row of Fig.~\ref{fig:Posterior distribution of source amplitude} show that the posterior distributions remain biased due to the persistent impact of model discrepancy on parameter inference. Next, we implement our proposed approach, in which the model is updated using the collected data, and the posterior distribution is subsequently re-evaluated. As illustrated in the second row of Fig.~\ref{fig:Posterior distribution of source amplitude}, the posterior distributions progressively converge toward the true parameter values and become more concentrated, indicating that the model correction effectively enhances parameter inference. We further check the performance of the updated network. The comparison of true $m^\dagger$, initial wrong $m$, and the corrected $m+\text{NN}$ is shown in Fig.~\ref{fig:Network-corrected model performance}. Within the data region, the network-corrected model $m+\text{NN}$ aligns much more closely with the true model $m^\dagger$, thereby demonstrating the informativeness of the data optimized for 
$A$ and the effectiveness of the correction approach. This improved model accuracy further enhances design optimization and parameter inference, accounting for the fine results in Fig.~\ref{fig:Posterior distribution of source amplitude}.

\end{document}